\documentclass[a4paper,fleqn,final]{cas-sc}

\usepackage[authoryear,longnamesfirst]{natbib}

\usepackage{graphicx}  
\usepackage[utf8]{inputenc} 
\usepackage[T1]{fontenc}    
\usepackage{hyperref}       
\usepackage{url}            
\usepackage{booktabs}       
\usepackage{multirow}  
\usepackage{amsfonts}       
\usepackage{nicefrac}       
\usepackage{microtype}      
\usepackage{xcolor}         
\usepackage{amsmath}
\usepackage{subcaption}
\usepackage{enumitem}
\usepackage{algorithm}
\usepackage{algorithmicx}
\usepackage{algpseudocode}
\usepackage{tabularx}
\usepackage{tcolorbox}
\usepackage{float}          
\usepackage{placeins}       

\begin{document}

\title{Latent Generative Modeling of Random Fields from Limited Training Data}

\shorttitle{Latent Generative Modeling of Random Fields}
\shortauthors{J. E. Warner et~al.}

\author[1]{James E. Warner}
\cormark[1]
\author[2]{Tristan A. Shah}
\author[1]{Patrick E. Leser}
\author[1]{Geoffrey F. Bomarito}
\author[3]{Joshua D. Pribe}
\author[3]{Michael C. Stanley}

\affiliation[1]{organization={NASA Langley Research Center},
                city={Hampton},
                postcode={23681}, 
                state={Virginia},
                country={USA}}
                
\affiliation[2]{organization={Texas Tech. University},
                city={Lubbock},
                postcode={79409}, 
                state={Texas},
                country={USA}}

\affiliation[3]{organization={Analytical Mechanics Associates},
                city={Hampton},
                postcode={23681}, 
                state={Virginia},
                country={USA}}

\cortext[cor1]{Corresponding author}

\begin{abstract}
The ability to accurately model random fields plays a critical role in science and engineering for problems involving uncertain, spatially-varying quantities such as heterogeneous material properties and turbulent flows. Deep generative models offer a powerful tool for sampling high- or infinite-dimensional uncertainties like random fields, but their reliance on large, dense training datasets limits their applicability in contexts where sufficient data is difficult or expensive to obtain. In this work, we propose a latent-space approach to generative modeling of random fields that incorporates domain knowledge to supplement limited training data. A constraint-aware variational autoencoder (VAE) with a function decoder is first used to learn compact latent representations of continuous functions that adhere to known physical or statistical constraints, even when training data is sparse or indirect. Generative modeling is then performed in the learned latent space, decoupling constraint enforcement from the sampling process. This decoupling enables expressive multi-step generative methods to be deployed in data-limited settings where existing constrained multi-step approaches are not directly applicable. The richer latent distributions captured by the generative model also overcome limitations of standard VAEs, which rely on simple parametric priors and struggle to represent complex, multimodal, or heavy-tailed distributions over functions. Efficacy is demonstrated on two challenging applications: wind velocity field reconstruction from sparse sensors and material property inference from indirect measurements. Results show the effectiveness of incorporating domain knowledge constraints for data-limited problems and the improved sample quality and robustness of the latent generative modeling approach versus directly sampling a constrained VAE. 
  
\end{abstract}



\maketitle

\section{Introduction}

Deep generative models (DGMs) have gained mainstream popularity by powering large language models (LLMs) \citep{openai2024gpt4} and text-to-image generators \citep{rombach2022high}, owing success to hardware and software breakthroughs, and, importantly, proliferation of massive training datasets. For science and engineering applications, data can be difficult and expensive to obtain, and the quantities being modeled are often infinite-dimensional functions of space and/or time, further exacerbating the data challenges. In order to fully leverage the power of DGMs in these scenarios, more research is needed on approaches to supplement the training of these models using domain knowledge to offset the data limitations.

This paper considers the challenge of learning to sample from distributions of continuous functions in the absence of large, dense training datasets. These \textit{random fields} are ubiquitous in science and engineering for modeling randomness in physical processes that vary spatially and/or temporally. For example, random fields are used to model material properties in heterogeneous media \citep{ostoja1998random}, contaminant spread in ground water  \citep{guo_2019}, velocity fields in turbulent flows \citep{doi:10.1137/S0036141002409167}, and temperature profiles and rainfall intensities for climate patterns \citep{Guillot_2015}. Training data for DGMs in these settings can be limited in three (not necessarily mutually exclusive) ways: 1) only relatively few total training data points are available, 2) each measurement may only provide partial/sparse information, and 3) only indirect measurements (i.e., of a related random field) are possible.

Depending on data availability, two common scientific problems are the \textit{reconstruction} of random fields from sparse measurements and \textit{inference}\footnote{We use "inference" to refer to solving inverse problems involving random fields, and mainly refer to the "inference" stage of DGMs as "sampling" to avoid confusion.} of unobserved random fields from indirect and potentially limited number of data. There is a subtle, but important, distinction between these problems that aim to recover a distribution of functions representing \textit{aleatory} uncertainty (or intrinsic randomness) versus reconstruction/inference of a single deterministic function with \textit{epistemic} uncertainty (or lack of knowledge) about its true value.  For example, consider the application of materials characterization from indirect measurements. One may seek to characterize the epistemic uncertainty about a deterministic, but unknown, material property in \textit{one particular} part or instead, to describe the variability of material properties across an \textit{entire population} of parts (aleatory uncertainty). This work specifically targets the latter, using DGMs to capture aleatory uncertainty in reconstruction and inference problems where the true solution is a random field. 

As DGMs have advanced from early approaches like generative adversarial networks (GANs) \citep{goodfellow2014} to modern diffusion-based models \citep{rombach2022high}, there has been a parallel effort to adapt these methods for scientific and engineering applications. At a high-level, DGMs aim to transform a simple reference distribution in a latent space to a complex, unknown data distribution, which can enable sample-based representations of uncertainty. A central theme in this body of work is the incorporation of domain knowledge through constraints, which serve to regularize learning and compensate for limited or indirect data. Most commonly, these constraints take the form of differential equation residuals, as popularized by physics-informed neural networks (PINNs) \citep{RAISSI2019686}, though statistical constraints have also been explored \citep{wu2020enforcing}. Constraints may be enforced either during training - typically via soft penalty terms added to the learning objective - or during sampling through conditioning mechanisms or projections that enforce hard constraint satisfaction on standard DGMs trained with data only. In data-limited regimes common to scientific applications, training-time constraints are often essential.

The first attempts at constrained generative modeling considered \textit{single-step} generative models such as GANs and variational autoencoders (VAEs) \citep{Kingma2013AutoEncodingVB}, which transform latent noise into samples in a single forward pass of a deep neural network. These models are attractive for scientific applications due to their computational efficiency and the fact that they can naturally produce function-valued output samples, making it straightforward to impose physics or statistical constraints via automatic differentiation. Early work used constrained GAN architectures to model random fields in fluid dynamics \citep{wu2020enforcing,xie_2018}, while adversarial and variational formulations were later extended to handle reconstruction and inference from sparse or noisy observations \citep{Yang_2019,Daw_2021,yang2018physicsinformedgenerativeadversarialnetworks,rodrigo-bonet_2024}. Among these, constrained VAEs, e.g., physics-informed VAEs (PI-VAEs) \citep{Zhong_2023}, are particularly relevant, as they model random fields from limited data using training-time constraint enforcement and serve as an important building block and baseline for the present work. Despite these advantages, single-step models struggle to represent complex or multimodal distributions: GANs are prone to training instability and mode collapse \citep{saxena2021generative}, while VAEs often suffer from degraded sample quality due to restrictive latent priors.

\textit{Multi-step} generative methods, including score-based/diffusion models \citep{ho2020denoising,nichol2021improved} and flow-matching formulations \citep{lipman2022flow,tong2023improving}, address many shortcomings of single-step models by constructing samples through an iterative refinement process that yields significantly improved stability and expressiveness. These approaches have inspired several attempts to impose physical constraints, such as training-time enforcement of partial differential equation (PDE) structure \citep{bastek2025physicsinformeddiffusionmodels}, inference-time conditioning or projection for constraint satisfaction \citep{jacobsen2024cocogenphysicallyconsistentconditionedscorebased, christopher2024constrainedsynthesisprojecteddiffusion}, and physics-guided diffusion for reconstruction or inverse problems \citep{Shu_2023, Dasgupta_2025, holzschuh2024improving}. Domain knowledge integration is more challenging relative to single-step methods, however, where special treatment is needed to apply constraints to intermediate, noisy samples \citep{bastek2025physicsinformeddiffusionmodels,liang_chance-constrained_2025}. Additionally, multi-step approaches largely work in a discretized (pixelized) space rather than producing continuous function samples. One known workaround is to adopt a functional generative modeling perspective \citep{utkarsh_physics-constrained_2025}. Finally, to the authors' knowledge, existing multi-step methods are not applicable for  settings where data is sparse or indirect \textit{at train time}, requiring fully-observed training data.

In this work, we adopt a \textit{latent-space} approach, arguing that it is uniquely suited for constrained generative modeling in science and engineering problems with limited training data. Latent generative models, such as latent diffusion and latent flow matching (LFM), learn expressive distributions in a low-dimensional latent space rather than directly in the high-dimensional physical domain, enabling efficient sampling, improved stability, and flexible representation of complex, multimodal distributions. Latent-space DGMs have been explored for inverse problems and field reconstruction, often through latent optimization or consistency constraints applied during sampling \citep{song2024solvinginverseproblemslatent,huang2024diffusionpdegenerativepdesolvingpartial,NEURIPS2023_cd830afc,zampini_training-free_2025}, and for generating continuous spatiotemporal fields from sparse measurements \citep{du2024confild}. However, most existing approaches assume fully observed training data and enforce constraints only at sampling time, limiting their effectiveness in data-limited scientific settings.

Motivated by these limitations, we introduce an approach for latent generative modeling of random fields with the following key features:
\begin{itemize}[itemsep=1pt, parsep=0pt]
	\item Estimates unknown random fields, producing samples of continuous space/time functions and capturing aleatory uncertainty in science and engineering phenomena. 
	\item Remains effective even when only sparse or indirect data is available \textit{at train time}.
	\item Applies generally to reconstruction and inference problems using both statistical or physical domain knowledge constraints.
\end{itemize}
Our approach involves two steps: 1) train a modified VAE to learn latent representations of functions that are consistent with sparse or indirect observations and available domain knowledge, incorporated through physical or statistical constraint residuals in the VAE loss, 2) perform generative modeling in the learned latent space, enabling high-fidelity sampling of random fields without requiring the latent distribution to conform to a simple parametric prior. While we employ LFM as one realization of this framework, the proposed approach is general and compatible with other latent generative modeling choices (e.g., latent diffusion). To support continuous function-valued outputs, an expressive function decoder is introduced, allowing generated samples to be queried and differentiated at arbitrary spatial or temporal locations during both training and sampling. 

Relative to existing constrained multi-step DGMs, our approach enables application to data-limited regimes by explicitly decoupling the learning of a compact, constraint-informed representation from the generative modeling. This substantially simplifies constraint integration and avoids the need to enforce physical/statistical consistency across noisy intermediate sampling states, while still producing continuous function-valued samples. In contrast to prior single-step constrained DGMs, which rely on simple latent priors and often struggle to capture complex or multimodal distributions, our approach leverages a richer learned latent space to represent aleatory uncertainty with higher fidelity. We explicitly demonstrate these benefits through direct comparisons with constrained VAEs, showing improved sample quality and robustness in sparse-data settings. The proposed framework is validated on two challenging applications - wind velocity field reconstruction from sparse sensors and material property inference from limited indirect measurements - highlighting the practical advantages of constrained latent generative modeling of random fields in science and engineering.

\section{Background}\label{sec:background}

\subsection{Random Fields for Science and Engineering}

A random field is defined as the $n$-dimensional mapping, $\mathbf{U}: \mathcal{X} \times \Omega \to \mathbb{R}^{n}$, where $\mathcal{X} \subset \mathbb{R}^{d_x}$ is the physical domain representing spatial/temporal coordinates ($1 \leq d_x \leq 4$), and $\Omega$ is a sample space accounting for randomness in $\mathbf{U}$. For each fixed $\hat{\omega} \in \Omega$, $\mathbf{u}(\mathbf{x}) \equiv \mathbf{U}(\mathbf{x},\hat{\omega})$ defines a deterministic function\footnote{We largely follow the convention of denoting random quantities with uppercase letters and deterministic realizations of random quantities with lowercase letters. Vector-valued quantities are identified by bold fonts.} over $\mathcal{X}$, while each fixed $\hat{\mathbf{x}} \in \mathcal{X}$, $\mathbf{U}(\hat{\mathbf{x}}, \omega)$ represents a random variable in $\mathbb{R}^{n}$.  The ability to generate samples of random fields, $\mathbf{u}(\mathbf{x})$, that are consistent with available data, physically-admissible, and can be evaluated at arbitrary $\mathbf{x}$ is of great importance to science and engineering applications. For example, a heterogenous material property may be modeled as a random field input to a physics-based model where the samples, $\mathbf{u}(\mathbf{x})$, are drawn to facilitate uncertainty quantification via Monte Carlo simulation. 

A common challenge of applying DGMs in this context is sparse and limited training data. More concretely, let $\mathcal{T}(\mathbf{u}(\mathbf{x})) = \mathbf{y} \in \mathbb{R}^{m \times n}$ be a measurement operator where $\mathbf{y} = \{\mathbf{u}(\mathbf{x}), \mathbf{x} \in \{\bar{\mathbf{x}}^{(k)} \}_{k=1}^m  \}$ are finite dimensional observations of $\mathbf{u}(\mathbf{x})$ and $\{\bar{\mathbf{x}}^{(k)} \}_{k=1}^m$ is a set of $m$ spatiotemporal coordinates where measurements are available. Assume a collection of $N$ independent observations are available for training, represented by the empirical probability distribution, $p(\mathbf{y} ) \equiv \{ \mathbf{y}^{(i)} \}_{i=1}^N$. There are many practical situations where $m$ is small (e.g., sparse measurements in space and/or time when few sensors are available) or $N$ is small (e.g., measurements are expensive and/or time consuming). Furthermore, many applications only allow \textit{indirect} observations, i.e., measurements of $\mathbf{U}$ are leveraged to perform inference on a related random field of interest, $\mathbf{V}(\mathbf{x}, \omega)$.  We focus on the challenges above but do not address \textit{noisy} measurements, an important limitation left to future work.

While science and engineering contend with smaller datasets than language and image applications, domain knowledge in the form of statistical and physical constraints can often be exploited, i.e.,
\begin{align}
	&R\left( \mathbf{U}(\mathbf{x}, \omega) \right) = 0, && \text{(Statistical Constraint)} \label{eq:statistical_constraint} \\
	&F\left( \mathbf{u}(\mathbf{x}), \mathbf{v}(\mathbf{x}) \right) = 0, && \text{(Physical Constraint)} \label{eq:physical_constraint}
\end{align}
where $\mathbf{v}(\mathbf{x}) \equiv \mathbf{V}(\mathbf{x},\hat{\omega})$.\footnote{Note that in the remainder of the paper, we will largely suppress the explicit $\mathbf{x}$ (and $\omega$) dependence of functions (and random fields) when they appear in operators and probability density functions for notational simplicity, e.g., $F(\mathbf{u}, \mathbf{v}) \equiv F\left( \mathbf{u}(\mathbf{x}), \mathbf{v}(\mathbf{x}) \right)$.}  $R$ enforces a known probabilistic metric and operates on random variables, e.g., prescribing a known mean, $\mu_u$, as $R\left( \mathbf{U} \right) = \mathbb{E}[\mathbf{U}] - \mu_u$.  $F$ typically represents the residual of known partial/ordinary differential equations that operate on deterministic functions, e.g., $F(\mathbf{u}, \mathbf{v}) = \mathcal{N}_x(\mathbf{u},  \mathbf{v}) $, where $\mathcal{N}_x$ is a differential operator. Two data-limited science and engineering problems demonstrating the use of these constraints are described next.

\subsection{Problems of Interest}

\vspace{1em} 
\begin{tcolorbox}[colback=blue!3!white, colframe=blue!75!black, arc=4pt, boxrule=1pt, left=5pt, right=5pt, top=5pt, bottom=5pt]
\subsubsection{Random Field Reconstruction from Sparse Data}\label{sec:rf_reconstruction}

\begin{itemize}[itemsep=1pt, parsep=1pt]
	\item \textbf{Goal:} Learn the probability distribution, $p(\mathbf{u})$, over functions, $\mathbf{u}(\mathbf{x})$, given the observations, $\{ \mathbf{y}^{(i)} \}_{i=1}^N$, and statistical constraint(s), $R(\mathbf{U})$. A DGM is sought that generates samples of functions, $\mathbf{u}(\mathbf{x})$, that satisfy $R$ and are consistent with observations.
	\item  \textbf{Data Characteristics:} There are sufficient numbers of observations ($N >> 1$) but each observation only provides sparse information on $\mathbf{u}(\mathbf{x})$, i.e., $m$ is small.
	\item \textbf{Target Application:} Wind velocity field estimation where probabilistic assessments of wind flows are required continuously over a spatial region and short time horizon, but wind measurements are only available on a relatively coarse spatiaotemporal grid.
\end{itemize}
\end{tcolorbox}

\vspace{0.5em} 

\begin{tcolorbox}[colback=green!3!white, colframe=green!75!black, arc=4pt, boxrule=1pt, left=5pt, right=5pt, top=5pt, bottom=5pt]
\subsubsection{Random Field Inference from Indirect Data}\label{sec:rf_inference}

\begin{itemize}[itemsep=1pt, parsep=1pt]
	\item \textbf{Goal:} Learn the joint probability distribution, $p(\mathbf{u}, \mathbf{v})$, over functions $\mathbf{u}(\mathbf{x})$ and $\mathbf{v}(\mathbf{x})$ given observations, $\{ \mathbf{y}^{(i)} \}_{i=1}^N$, and physical constraints relating the functions, $F(\mathbf{u}, \mathbf{v})$. A DGM is sought that generates pairs of functions, ($\mathbf{u}(\mathbf{x})$, $\mathbf{v}(\mathbf{x})$), that satisfy $F$ and where samples, $\mathbf{u}(\mathbf{x})$, are consistent with observations.
	\item  \textbf{Data Characteristics:} The primary quantity of interest, $\mathbf{v}(\mathbf{x})$, is unobservable. A related quantity, $\mathbf{u}(\mathbf{x})$, is observed with sufficient numbers of sensors per observation ($m >> 1$) but the total number of observations may be limited, i.e., $N$ may be small.
	\item \textbf{Target Application:} Material property characterization where the property varies spatially within a test article and also randomly between different test articles, e.g., due to manufacturing variability. Experimental techniques provide dense measurements of displacement fields per test but only relatively few tests can be performed.
\end{itemize}
\end{tcolorbox}

\subsection{Random Field Vs. Deterministic Function Estimation}

There is an important distinction between the reconstruction and inference of random fields (the focus of this work) versus deterministic functions (the focus of many existing works \citep[e.g.,][]{Yang_2019,yang2019highlyscalablephysicsinformedganslearning,Shu_2023}). For random fields, we leverage DGMs to model the \textit{aleatory} uncertainty (intrinsic randomness) as a probability distribution over deterministic functions. In this case, additional data provides more precise estimates of the uncertainty but does not reduce it. When estimating a deterministic function, DGMs quantify \textit{epistemic} uncertainty (lack of knowledge about the true function) due to limited and/or noisy data. Here, additional data reduces uncertainty in the estimate of the deterministic function. Mathematically, most previous work has approximated conditional probability distributions for both reconstruction, $p(\mathbf{u} | \mathcal{T}(\mathbf{u}) = \mathbf{y}^{(i)})$ \citep{du2024confild, tran2020gans}, and inference,  $p(\mathbf{v}  | \mathcal{T}(\mathbf{u}) = \mathbf{y}^{(i)})$ \citep{Dasgupta_2025, NEURIPS2023_cd830afc}, while we seek the marginal/joint distributions $p(\mathbf{u})$ and $p(\mathbf{u}, \mathbf{v})$.

\subsection{Summary of Related Work}

This section briefly summarizes  existing work on constrained DGMs for science and engineering. To situate the proposed approach within this research area, Table \ref{tab:related_work_table} provides a comparative assessment of select related work. This table classifies methods based on the type of solutions they target (random fields vs. deterministic functions), their ability to handle limited training data and produce continuous samples over space and time, the nature of the  constraints they apply, and which DGM type/architecture they adopt.

From the table, several trends and observations emerge. While multi-step and latent-space approaches represent current state-of-the-art for DGMs, the authors are not aware of any existing methods that enable these architectures to be applied when data is limited (sparse/indirect) at train time. These approaches have also largely operated in a discrete (pixelized) output space, with one notable exception being recent work based on functional flow matching \citep{utkarsh_physics-constrained_2025}. Single-step methods, despite issues such as  training instability and sample quality, are better equipped to produce continuous samples of random fields from limited training data, making them more relevant for the problems addressed in this work. 

The proposed work highlights the use of a latent-space approach as a means of enabling more expressive multi-step DGMs to remain effective in limited training data regimes. To the authors' knowledge, this work is also the first to demonstrate the generality of a single method to leverage both physical and statistical constraints to solve random field inference and reconstruction problems. Of the relatively few methods in  Table \ref{tab:related_work_table} that are applicable to the problems of interest in this study, a constrained VAE approach, similar to the PI-VAE of \cite{Zhong_2023}, is adopted as a baseline since it is most similar to the proposed method.

\setlength{\extrarowheight}{3pt} 

\begin{table}[ht]
\centering
\renewcommand{\arraystretch}{1.1} 
\setlength{\tabcolsep}{5pt} 

\begin{tabularx}{\textwidth}{>{\raggedright\arraybackslash}p{0.155\textwidth}>{\raggedright\arraybackslash}p{0.12\textwidth}>{\raggedright\arraybackslash}p{0.1\textwidth}>{\raggedright\arraybackslash}p{0.09\textwidth}>{\raggedright\arraybackslash}p{0.08\textwidth}>{\raggedright\arraybackslash}p{0.11\textwidth}>{\raggedright\arraybackslash}p{0.09\textwidth}X}
\hline
\textbf{Work} & \textbf{Solution Type} & \textbf{Limited Train Data?} & \textbf{Continuous Samples?} & \textbf{Constraint Type} & \textbf{DGM Type} & \textbf{Constraint Timing} & \textbf{DGM} \\
\hline
\textbf{Our Approach} & \textbf{Random Field} & \textbf{Yes} & \textbf{Yes} & \textbf{Both} & \textbf{Latent Space} & \textbf{Training} & \textbf{LFM} \\
Zhong (2023) & Random Field & Yes & Yes & Physical & Single Step & Training & VAE \\
Yang (2019) & Random Field & Yes & Yes & Physical & Single Step & Training & GAN \\
Wu (2020) & Random Field & Yes* & Yes & Statistical & Single Step & Training & GAN \\
Daw (2021) & Det. Function & Yes & Yes & Physical & Single Step & Training & GAN \\
Yang (2019) & Det. Function & Yes & Yes & Physical & Single Step & Training & GAN \\
Bastek (2025) & Random Field & No & No & Physical & Multi Step & Training & DM \\
Jacobsen (2024) & Both & No & No & Physical & Multi Step & Sampling & DM \\
Christopher (2024) & Random Field & No & No & Physical & Multi Step & Sampling & DM \\
Huang (2024) & Det. Function & No & No & Physical & Multi Step & Sampling & DM \\
Holzschuh (2024) & Det. Function & No & No & Physical & Multi Step & Sampling & FM \\
Liang (2025) & Random Field & No & No & Physical & Multi Step & Sampling & FM \\
Utkarsh (2025) & Random Field & No & Yes & Physical & Multi Step & Sampling & FFM \\
Song (2024) & Det. Function & No & No & Physical & Latent Space & Sampling & LDM \\
Shmakov (2023) & Det. Function & No & No & Physical & Latent Space & Training & LDM \\
Zampini (2025) & Random Field & No & No & Physical & Latent Space & Sampling & LDM \\
\hline
\end{tabularx}
\caption{Summary of select related work on constrained DGMs for science and engineering compared to the proposed approach. \\ * indicates the approach can theoretically handle limited train data but this was not demonstrated. }
\label{tab:related_work_table}
\end{table}

\section{Methodology}

\begin{figure}[htbp]
    \centering
    \includegraphics[width=\textwidth]{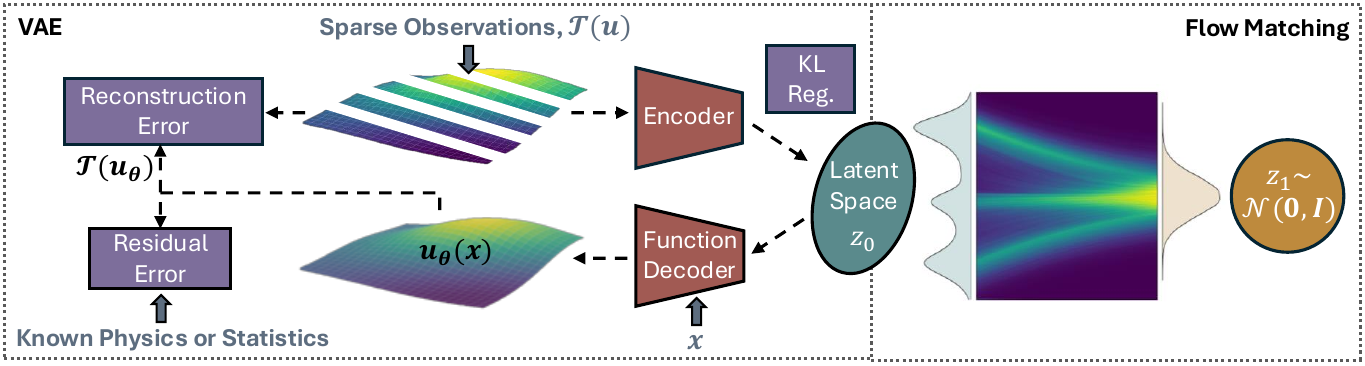}
    \caption{\textbf{Latent generative modeling of random fields from limited data with c-LFM}. A constraint-aware VAE with function decoder learns a latent representation of the continuous random field, $\mathbf{U}(\mathbf{x}, \omega)$, from sparse observations. VAE loss terms are shown in purple boxes, where a residual error supplements limited data with statistical or physical constraints. Flow matching enables sampling from a potentially complex latent distribution.}
    \label{fig:figure1}
\end{figure}

An overview of the framework is shown in Figure \ref{fig:figure1}. The foundation of our approach involves generative modeling in the latent space of a pretrained VAE. The VAE encodes discrete (potentially sparse) data into latent variables that are then transformed into continuous function samples using a function decoder. Known physical and statistical constraints are incorporated into the VAE loss function through a residual error. The latent space provides a compressed (typically low-dimensional) function parameterization that lends itself to straightforward latent generative modeling. In this work, we demonstrate flow matching for sampling the VAE latent space as one instantiation of the framework, which we refer to as constrained LFM (c-LFM).  

Training c-LFM is a two-stage process: 1) train a constraint-aware VAE from limited training data, 2) train flow matching to generate latent vectors of the  pretrained VAE. Random field samples are then generated by sampling latent vectors with flow matching and transforming them using the VAE function decoder. These processes will be expanded upon in the following subsections; further details can be found in Algorithms \ref{alg:vaeTrain}--\ref{alg:lfmSample}.


\subsection{VAE with Function Decoder and Constraints}\label{sec:vae}

VAEs are single-step generative models that use latent variable representations to probabilistically model complex data distributions. In this work, a VAE is used to learn a compact latent representation of functions, which can be subsequently randomly sampled and transformed into samples of a random field of interest. We propose a modified VAE in which the learned latent space is mapped to continuous function samples that satisfy prescribed constraints and are informed by (limited) available training data.

Following the seminal work on VAEs \citep{Kingma2013AutoEncodingVB}, it is assumed that samples, $\mathbf{u}(\mathbf{x})$, from the random field, $\mathbf{U}$, are generated from the conditional distribution, $p(\mathbf{u} | \mathbf{z}) $, where $\mathbf{z} \in \mathbb{R}^{d_z}$ are $d_z$ unobserved latent variables. Here, $\mathbf{z}$ can be viewed as accounting for stochasticity in the random field, i.e., $\mathbf{U}(\mathbf{x}, \omega) = \mathbf{U}(\mathbf{x}, \mathbf{Z}(\omega))$.  We further assume that the latent variables are described by a standard normal prior distribution, $p(\mathbf{z}) =  \mathcal{N}(\mathbf{0}, \mathbf{I})$. The posterior distribution, $q(\mathbf{z} | \mathbf{y})$, then describes the distribution of $\mathbf{z}$ conditioned on discrete, and typically limited, observations, $\mathbf{y} = \mathcal{T}(\mathbf{u})$ (Section \ref{sec:background}). Our approach thus requires a mechanism to sample both $\mathbf{z}^{(j)} \sim q(\mathbf{z} | \mathbf{y})$ and $\mathbf{u}^{(j)} \sim p(\mathbf{u} | \mathbf{z} = \mathbf{z}^{(j)})$.

To this end, VAEs approximate the conditional probability distributions, $q(\mathbf{z} | \mathbf{y})$ and $p(\mathbf{u} | \mathbf{z})$, with neural networks termed the encoder and decoder, respectively. The encoder network, $\mathcal{E}_{\boldsymbol{\phi}}$, is parametrized by $\boldsymbol{\phi}$ and maps discrete observations to the latent space by outputting the mean and standard deviation vectors of a multivariate normal distribution, $(\boldsymbol{\mu}_{\boldsymbol{\phi}},  \boldsymbol{\sigma}_{\boldsymbol{\phi}} ) = \mathcal{E}_{\boldsymbol{\phi}}(\mathbf{y})$.  Latent variables are subsequently sampled from a variational approximation to the posterior, $q_{\boldsymbol{\phi}}(\mathbf{z} | \mathbf{y}) = \mathcal{N}(\boldsymbol{\mu}_{\boldsymbol{\phi}}, \mathrm{diag}(\boldsymbol{\sigma}^2_{\boldsymbol{\phi}}))$. To ensure gradients with respect to encoder parameters are differentiable during training, the reparameterization trick \citep{Kingma2013AutoEncodingVB} is employed to separate the  randomness, $\epsilon \sim \mathcal{N}(\mathbf{0}, \mathbf{I}_{d_z})$, from the learned parameters, i.e., $\mathbf{z} = \boldsymbol{\mu}_{\boldsymbol{\phi}} + \epsilon\ \circ \ \boldsymbol{\sigma}_\phi$ where $\circ$ denotes elementwise multiplication.

To decode a latent sample, $\mathbf{z}^{(j)}$, into a continuous function, $\mathbf{u}^{(j)}(\mathbf{x}) \sim p_{\boldsymbol{\theta}}(\mathbf{u} | \mathbf{z} = \mathbf{z}^{(j)})$, our first modification to the standard VAE is to implement a function decoder, $\mathcal{D}_{\boldsymbol{\theta}}$, using the DeepONet \citep{lu2021learning} architecture:
\begin{equation}
    \mathcal{D}_{\boldsymbol{\theta}}^u(\mathbf{z}, \mathbf{x}) \equiv \sum_{k=1}^p b^u_k(\mathbf{z};\boldsymbol{\psi}) t^u_k(\mathbf{x}; \boldsymbol{\xi}). \label{eq:decoder}
\end{equation}
Here, $\boldsymbol{\theta} = [\boldsymbol{\psi}, \boldsymbol{\xi}]$, and $p$ is a hyperparameter controlling the number of learned basis functions. The  decoder output is the dot product of a \textit{branch} network with parameters $\boldsymbol{\psi}$: $\mathbf{b}^u: \mathbf{R}^{d_z}  \to  \mathbf{R}^{p}$, which receives latent variables as inputs, and a \textit{trunk} network with parameters $\boldsymbol{\xi}$: $\mathbf{t}^u: \mathbf{R}^{d_x} \to  \mathbf{R}^{p}$, which receives spatial/temporal coordinates as inputs. This architecture decomposes $\mathbf{U}(\mathbf{x}, \omega)$ into stochastic coefficients and deterministic basis functions, as is common in random field modeling \citep{karhunen1947lineare,loeve1955probability,xiu2010numerical}, and provides a simple and expressive method of evaluating samples arbitrarily in $\mathbb{R}^{d_x}$. Note that in principle our framework should be compatible with other function parameterizations as well. For the case of random field inference (Section  \ref{sec:rf_inference}), a second decoder for $\mathbf{v}(\mathbf{x})$, $ \mathcal{D}_{\boldsymbol{\theta}}^v(\mathbf{z}, \mathbf{x})$, is defined similarly.


The encoder and decoder networks are typically trained by optimizing the evidence lower bound (ELBO) loss, which balances two objectives: reconstructing the observations, $\mathbf{y}$, and regularizing the latent variable distribution. For a single batch of data, $\{ \mathbf{y}^{(j)} \}$, the ELBO loss is defined as 
\begin{equation}
	\mathcal{L}_{ELBO}(\boldsymbol{\phi}, \boldsymbol{\theta}; \mathbf{y}^{(j)}) = \| \mathcal{T}( \mathcal{D}_{\boldsymbol{\theta}}^u(\mathbf{z}^{(j)} , \mathbf{x})) - \mathbf{y}^{(j)}  \|^2 + \lambda_{kl} D_{KL} \left(q_{\boldsymbol{\phi}}(\mathbf{z}^{(j)}  | \mathbf{y}^{(j)} ), p(\mathbf{z}) \right),  \label{eq:elbo} 
\end{equation}
where  $D_{KL}$ is the Kullback-Leibler (KL) divergence, which penalizes deviations of the approximate posterior from the prior, and $ \lambda_{kl}$ is a weight that controls the relative importance of this prior regularization term. The ELBO loss thus aims to learn latent representations that not only reconstruct the data accurately but also enable sampling through effective posterior regularization. Once trained, the VAE can be used as a generative model by randomly sampling the prior and evaluating the decoder, i.e., $\mathbf{u}^{(j)}(\mathbf{x}) = \mathcal{D}_{\boldsymbol{\theta}}^u(\mathbf{z}^{(j)}, \mathbf{x})$ with $\mathbf{z}^{(j)} \sim \mathcal{N}(\mathbf{0}, \mathbf{I})$ (the encoder is no longer needed).

When training data is limited, optimizing the ELBO loss in Equation \eqref{eq:elbo} directly is not generally feasible. For example, in random field reconstruction problems (Section \ref{sec:rf_reconstruction}) where $\mathbf{y}$ only provides sparse measurements of $\mathbf{U}(\mathbf{x}, \omega)$, there is no mechanism to learn the behavior of the random field away from measurement locations. Likewise, for inference problems (Section \ref{sec:rf_inference}), the ELBO loss has no dependence on the unobserved random field of interest, $\mathbf{V}(\mathbf{x}, \omega)$. To make VAE training tractable for these settings, we update the standard ELBO loss to incorporate statistical and physical constraints as follows:
\begin{equation}
    \begin{aligned}
        \mathcal{L}(\boldsymbol{\phi}, \boldsymbol{\theta}; \mathbf{y}^{(j)} ) & = \mathcal{L}_{ELBO}(\boldsymbol{\phi}, \boldsymbol{\theta}; \mathbf{y}^{(j)}) +  \lambda_r  \| R( \mathcal{D}_{\boldsymbol{\theta}}^u(\mathbf{z}^{(j)}, \mathbf{x}))   \|^2  \quad + \lambda_f \| F( \mathcal{D}_{\boldsymbol{\theta}}^u(\mathbf{z}^{(j)}, \mathbf{x}),  \mathcal{D}_{\boldsymbol{\theta}}^v(\mathbf{z}^{(j)}, \mathbf{x}))   \|^2 \\
                                                                               & =\underbrace{\| \mathcal{T}( \mathcal{D}_{\boldsymbol{\theta}}^u(\mathbf{z}^{(j)} , \mathbf{x})) - \mathbf{y}^{(j)}  \|^2}_{\text{Reconstruction error}}
        \quad + \underbrace{\lambda_{kl} D_{KL} \left(q_{\boldsymbol{\phi}}(\mathbf{z}^{(j)}  | \mathbf{y}^{(j)} ), p(\mathbf{z}) \right)}_{\text{KL regularization}}                                                                                                                                                                                                                                                                \\
                                                                               & \quad + \underbrace{\lambda_r  \| R( \mathcal{D}_{\boldsymbol{\theta}}^u(\mathbf{z}^{(j)}, \mathbf{x}))   \|^2}_{\text{Statistics residual}}
        \quad + \underbrace{\lambda_f \| F( \mathcal{D}_{\boldsymbol{\theta}}^u(\mathbf{z}^{(j)}, \mathbf{x}),  \mathcal{D}_{\boldsymbol{\theta}}^v(\mathbf{z}^{(j)}, \mathbf{x}))   \|^2}_{\text{Physics residual}}.   \label{eq:vae_loss}
    \end{aligned}
\end{equation}
Here, the statistics and physics residuals quantify the violation of the constraints in Equations \eqref{eq:statistical_constraint} and \eqref{eq:physical_constraint} when evaluated with the decoder approximations to $\mathbf{u}(\mathbf{x})$ and $\mathbf{v}(\mathbf{x})$, enabling learning in data-limited regimes. The weights, $\lambda_r$ and  $\lambda_f$, control the relative importance of the statistics and physics residuals, respectively. The expected loss, $\mathbb{E}[ \mathcal{L}] = \frac{1}{B} \sum_{j=1}^{B}  \mathcal{L}(\boldsymbol{\phi}, \boldsymbol{\theta}; \mathbf{y}^{(j)} )$, is minimized during training with $B$ batches.

\subsubsection{Evaluating Constraint Residuals}\label{sec:constraints}

The statistics and physics residuals in Equation \ref{eq:vae_loss} must be estimated at each optimization iteration while training the VAE. Since the constraints in Equations \ref{eq:statistical_constraint} and \ref{eq:physical_constraint} are intended to hold for all $\mathbf{x}$, we adopt a common approach of estimating the residual terms over a finite set of collocation points $\{ \tilde{\mathbf{x}}^{(c)} \}_{c=1}^{C}$ \citep{RAISSI2019686}. We choose these points randomly at each training iteration, $ \tilde{\mathbf{x}}^{(c)} \sim \text{Uniform}(\mathcal{X})$.

The manner in which the residuals are computed varies between statistical and physical constraints. For the former, any statistics appearing in $R$ must be estimated using the current batch of samples during training,
\begin{equation}
          \mathbb{E}_{\mathbf{Z}} \left[R(U(\mathbf{x}, \mathbf{Z}(\omega)))\right]  \approx \frac{1}{N_B} \sum_{i=1}^{N_B}  \left[ R( \mathcal{D}_{\theta}^u(\mathbf{z}^{(j, i)}, \mathbf{x}) )\right] = 0 
\end{equation}
where $N_B = \frac{N}{B}$ is the number of samples per batch and $\mathbf{z}^{(j, i)}$ denotes the $i^{th}$ latent variable sample in batch $j$ where $\mathbf{z}^{(j)} \sim q_{\boldsymbol{\phi}}(\mathbf{z} | \mathbf{y}^{(j)})$.  Thus, the residual of the statistical constraint is evaluated at the set of collocation points as follows:
\begin{equation}
    \| R( \mathcal{D}_{\theta}^u(\mathbf{z}^{(j)}, \mathbf{x}))   \|^2  \approx \frac{1}{C} \sum_{c=1}^C \left\| \frac{1}{N_B} \sum_{i=1}^{N_B}  \left[ R( \mathcal{D}_{\theta}^u(\mathbf{z}^{(j, i)},  \tilde{\mathbf{x}}^{(c)} ) )\right] \right\|^2.
\end{equation}

On the other hand, the physical constraints are intended to hold both for all $\mathbf{x}$ and for all samples of $\mathbf{u}$ and $\mathbf{v}$. Thus, the residuals are evaluated on a per-latent-variable-sample basis as follows:

\begin{equation}
    \| F( \mathcal{D}_{\theta}^u(\mathbf{z}^{(j)}, \mathbf{x}),  \mathcal{D}_{\theta}^v(\mathbf{z}^{(j)}, \mathbf{x}))   \|^2  \approx \frac{1}{C} \frac{1}{N_B} \sum_{c=1}^C  \sum_{i=1}^{N_B}  \left\| F( \mathcal{D}_{\theta}^u(\mathbf{z}^{(j, i)},  \tilde{\mathbf{x}}^{(c)}),  \mathcal{D}_{\theta}^v(\mathbf{z}^{(j, i)}, \tilde{\mathbf{x}}^{(c)}))   \right\|^2,
\end{equation}
where any needed derivatives for the evaluation of $F$ are implemented with autodifferentiation.

A summary of the complete VAE training procedure is provided in Algorithm \ref{alg:vaeTrain}. Note that simultaneous statistical and physical constraints is possible; however, we apply only one at a time (statistical for reconstruction and physical for inference) in this work.

\begin{algorithm}[H]
    \caption{Training VAE with Function Decoder and Residual Constraints}
    \begin{algorithmic}[1]
        \Require Training data $\{\mathbf{y}^{(j)}\}_{j=1}^{N}$, encoder $\mathcal{E}_{\phi}$, function decoder $\mathcal{D}_{\theta}$, weights $\lambda_{kl}$, $\lambda_r$, $\lambda_f$
        \For{each training iteration}
        \For{each batch $j = 1, ..., B$}
        \State Sample batch $\mathbf{y}^{(j)}$  and collocation points $\{\tilde{\mathbf{x}}^{(c)}\}_{c=1}^C$
        \State $\mathbf{\mu_{\phi}}, \mathbf{\sigma_{\phi}} = \mathcal{E}_{\phi}(\mathbf{y}^{(j)})$
        \State $\mathbf{z}^{(j)} = \mu_{\phi} + \epsilon \ \circ \ \sigma_{\phi}$, with $\epsilon \sim \mathcal{N}(\mathbf{0}, \mathbf{I}_{d_z})$
        \State Compute reconstruction: $\hat{\mathbf{y}}^{(j)} = \mathcal{T}(\mathcal{D}_{\theta}^u(\mathbf{z}^{(j)}, \mathbf{x}))$
        \State $\mathcal{L}_{rec} \mathrel{+}= \|\hat{\mathbf{y}}^{(j)} - \mathbf{y}^{(j)}\|^2$
        \State $\mathcal{L}_{kl} \mathrel{+}= D_{KL}(q_{\phi}(\mathbf{z}^{(j)}|\mathbf{y}^{(j)}), p(\mathbf{z}))$
        \State $\mathcal{L}_{stat} \mathrel{+}= \| R( \mathcal{D}_{\theta}^u(\mathbf{z}^{(j)}, \tilde{\mathbf{x}}^{(c)}))   \|^2$
        \State $\mathcal{L}_{phys} \mathrel{+}= \| F( \mathcal{D}_{\theta}^u(\mathbf{z}^{(j)}, \tilde{\mathbf{x}}^{(c)}),  \mathcal{D}_{\theta}^v(\mathbf{z}^{(j)}, \tilde{\mathbf{x}}^{(c)}))   \|^2$
        \EndFor
        \State $\mathcal{L} = \mathcal{L}_{rec} + \lambda_{kl}\mathcal{L}_{kl} + \lambda_r\mathcal{L}_{stat} + \lambda_f\mathcal{L}_{phys}$
        \State Update parameters $\boldsymbol{\phi}, \boldsymbol{\theta}$ to minimize $\mathcal{L}$
        \EndFor
        \Ensure Trained encoder $\mathcal{E}_{\phi}$ and decoder $\mathcal{D}_{\theta}$
    \end{algorithmic}
    \label{alg:vaeTrain}
\end{algorithm}

%
%

\subsection{Latent Generative Modeling with Flow Matching}\label{sec:latent_flow_matching}

After training the VAE, generative modeling is performed in the learned latent space to draw compressed representations of functions, which can then be decoded into continuous random field samples. Here, a generative model is trained to sample latent vectors from the aggregate (marginal) posterior,
\begin{equation}
q_{\boldsymbol{\phi}}(\mathbf{z}) \approx \frac{1}{N} \sum_{i=1}^{N} q_{\boldsymbol{\phi}}(\mathbf{z} | \mathbf{y}^{(i)}), 
\end{equation}
which represents the distribution of encoded training data, rather than simply sampling the prior $p(\mathbf{z})$ as with standard VAEs. Latent space generative modeling techniques originated as a computationally efficient and scalable solution to overcome the limitations of multi-step methods that traditionally operated in high-dimensional data spaces (e.g., pixel space). These approaches require minimal regularization of the latent space during VAE training ($\lambda_{kl} \approx 10^{-6}$), enabling the VAE to primarily focus on producing high-fidelity reconstructions of the data, and in our case, ensuring adherence to physical and statistical constraints. The latent generative model effectively samples the latent space, addressing the increased complexity introduced by relaxed regularization.

In this work, we adopt latent flow matching for generative modeling in the latent space but note that other methods are equally valid within our framework. Let $\mathbf{z}_0 \sim q_{\boldsymbol{\phi}}(\mathbf{z} | \mathbf{y})$ be an encoded training observation\footnote{Recall that samples can be produced from $q_{\boldsymbol{\phi}}(\mathbf{z} | \mathbf{y}=\mathbf{y}^{(j)})$ by first encoding via $\mathcal{E}_{\phi}(\mathbf{y}^{(j)})$ and then applying the reparameterization trick as outlined in Section \ref{sec:vae}. This requires $\mathbf{y}^{(j)}$ to be available, which is true during training.} and $\mathbf{z}_1\sim \mathcal{N}(\mathbf{0}, \mathbf{I})$ denote a sample from random noise.  Flow matching posits the following ordinary differential equation (ODE) to transform samples from the source (latent) distribution to reference (noise) distribution:
\begin{equation}
    \frac{d \mathbf{z}_t}{d t} = \boldsymbol{\nu}(\mathbf{z}_t, t) \label{eq:flow_matching_ode}
\end{equation}
where time $t \in [0, 1]$ and $\boldsymbol{\nu}: \mathbb{R}^{d_z} \times [0, 1] \to \mathbb{R}^{d_z}$ is the time-dependent velocity field. Here, $t=0$ represents a clean latent vector sample and $t=1$ represents the reference (noise) state. After training a neural network approximation of the velocity, the generation process involves integrating the ODE backward in time to generate new latent samples from noise.

To facilitate training with flow matching, a parameterization for $\mathbf{z}_t$ is required. We choose simple linear interpolation between reference and source samples, $\mathbf{z}_t \equiv (1 - t) \mathbf{z}_0 + t \mathbf{z}_1$  (the \textit{optimal transport} map \citep{lipman2022flow}), which is a common choice that yields a constant velocity ODE with $\boldsymbol{\nu} =  \mathbf{z}_1 -  \mathbf{z}_0$. The neural network approximation for velocity, $\boldsymbol{\nu}_{\eta}$ with parameters $\eta$, is then introduced and trained using
\begin{align}
    \eta^* &= \underset{\eta}{\arg\min} \; \mathbb{E}_{t, \mathbf{z}_t} \| \boldsymbol{\nu} - \boldsymbol{\nu}_{\eta}(\mathbf{z}_t, t) \|_2^2  \nonumber \\
    &=   \underset{\eta}{\arg\min} \; \mathbb{E}_{t, \mathbf{z}_t} \| \mathbf{z}_1 -  \mathbf{z}_0 - \boldsymbol{\nu}_{\eta}(\mathbf{z}_t, t) \|_2^2, \label{eq:flow_match_training}
\end{align}
where the expectation can be computed over samples $t^{(j)}\sim\text{Uniform}(0, 1)$ and $\mathbf{z}_t^{(j)}$, which is sampled by interpolating between $\mathbf{z}_0^{(j)}\sim q_{\boldsymbol{\phi}}(\mathbf{z} | \mathbf{y}^{(j)})$ and $\mathbf{z}_1^{(j)} \sim \mathcal{N}(\mathbf{0}, \mathbf{I})$ at the given $t^{(j)}$. Here $\mathbf{y}^{(j)} \sim p(\mathbf{y})$ represents the same data that was used to train the VAE via Algorithm \ref{alg:vaeTrain}.

Once flow matching training has been completed ($\boldsymbol{\nu}_{\eta^*}$ has been learned), realizations of $\mathbf{z}_0$ can be generated by numerically integrating Equation \ref{eq:flow_matching_ode} in reverse with random noise as the terminal condition and using $\boldsymbol{\nu}_{\eta^*}$ as the right-hand side. Here, we use a simple Euler method, i.e., $\mathbf{z}_{t-\Delta t} = \mathbf{z}_{t} - \Delta t \boldsymbol{\nu}_{\eta}(\mathbf{z}_t, t)$ where $\Delta t>0$ is the time step used for discretization. Resulting latent variables can then be decoded to produce continuous samples of $\mathbf{u}(\mathbf{x})$ and/or $\mathbf{v}(\mathbf{x})$. See Algorithms \ref{alg:lfmTrain} and \ref{alg:lfmSample} for more details on training and sampling, respectively.

\begin{algorithm}[H]
    \caption{Latent Flow Matching Training}
    \begin{algorithmic}[1]
        \Require Trained (fixed) encoder $\mathcal{E}_{\phi}$, velocity network $\boldsymbol{\nu}_{\eta}$
        \For{each training iteration}
        \For{each batch $j = 1, ..., B$}
        \State $\mathbf{y}^{(j)} \sim p(\mathbf{y})$
        \State $\mathbf{z}_0^{(j)}\sim q_{\boldsymbol{\phi}}(\mathbf{z} | \mathbf{y} =\mathbf{y}^{(j)})$,\ \ $\mathbf{z}_1^{(j)} \sim \mathcal{N}(\mathbf{0}, \mathbf{I})$
        \State $t^{(j)} \sim \text{Uniform}(0, 1)$
        \State $\mathbf{z}_t^{(j)} = (1 - t^{(j)}) \mathbf{z}_0^{(j)} + t^{(j)} \mathbf{z}_1^{(j)}$
        \State $\boldsymbol{\nu}^{(j)} = \mathbf{z}_1^{(j)} - \mathbf{z}_0^{(j)}$
        \State $\mathcal{L} \mathrel{+}=  \|\boldsymbol{\nu}^{(j)} - \boldsymbol{\nu}_{\eta}(\mathbf{z}_t^{(j)}, t^{(j)})\|_2^2$
        \EndFor
        \State Update parameters $\eta$ to minimize $\mathcal{L}$
        \EndFor
        \Ensure Trained velocity network $\boldsymbol{\nu}_{\eta}$
    \end{algorithmic}
    \label{alg:lfmTrain}
\end{algorithm}
\begin{algorithm}[H]
    \caption{Constrained Latent Flow Matching (c-LFM) Sampling}
    \begin{algorithmic}[1]
        \Require Trained velocity network $\boldsymbol{\nu}_{\eta^*}$, trained decoder $\mathcal{D}_{\theta}^u$
        \State $\mathbf{z}_1 \sim \mathcal{N}(\mathbf{0}, \mathbf{I})$
        \State Solve $\frac{d\mathbf{z}_t}{dt} = -\boldsymbol{\nu}_{\eta^*}(\mathbf{z}_t, t)$ from $t=1$ to $t=0$, starting from $\mathbf{z}_1$
        \State $\mathbf{u}(\mathbf{x}) = \mathcal{D}_{\theta}^u(\mathbf{z}_0, \mathbf{x}) = \sum_{k=1}^p b^u_k(\mathbf{z}_0; \psi) t^u_k(\mathbf{x}; \xi)$
        \Ensure Continuous function $\mathbf{u}(\mathbf{x})$
    \end{algorithmic}
    \label{alg:lfmSample}
\end{algorithm}

\subsection{Latent Space Sampling Comparison}\label{sec:sampling_approaches}

In order to highlight the advantages of our constrained latent-space DGM method versus a simpler single-step approach using the VAE directly, we will study the accuracy of decoded random field samples, $\mathbf{u}^{(j)}(\mathbf{x}) = \mathcal{D}_{\boldsymbol{\theta}}^u(\mathbf{z}^{(j)}, \mathbf{x})$, with three distinct latent sampling strategies for $\mathbf{z}^{(j)}$:

\begin{itemize}[itemsep=1pt, parsep=0pt]
	\item \textbf{Prior sampling (constrained VAE baseline)}: $\mathbf{z}^{(j)} \sim \mathcal{N}(\mathbf{0}, \mathbf{I})$ \\
This case represents generative sampling with standard VAEs where latent vectors are simply drawn from the fixed prior distribution. Here, a higher $\lambda_{KL}$ is required to enforce latent space regularization.
	\item \textbf{Posterior sampling (best-case)}: $\mathbf{z}^{(j)}\sim q_{\boldsymbol{\phi}}(\mathbf{z} \mid \mathbf{y} = \mathbf{y}^{(j)})$ \\
This case represents direct sampling from the learned posterior of the pre-trained VAE by encoding data to latent vectors. This serves as a best-case reconstruction benchmark but is not usable in practice because it requires test data, $\mathbf{y}^{(j)} \sim p(\mathbf{y})$, as an input.
	\item \textbf{LFM sampling (ours)}: $\mathbf{z}^{(j)}$ sampled according to lines 1 and 2 of Algorithm \ref{alg:lfmSample} \\
Our proposed approach approximates the true latent posterior of the VAE, sampling latent vectors using flow matching and avoiding potential overregularization of the learned latent representation (lower $\lambda_{KL}$).
\end{itemize}
We argue, and show in the examples, that latent space flexibility is required for effective data reconstruction and constraint satisfaction in many cases, which limits the performance of the constrained VAE baseline but allows the proposed c-LFM approach to excel, approaching best-case posterior sampling performance.

\begin{figure}[htbp]
    \centering
    \includegraphics[width=2.9in]{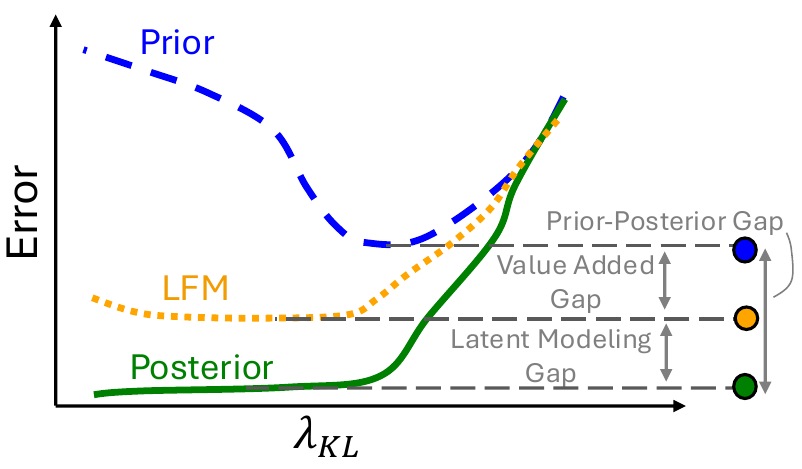}
    \caption{A notional diagram comparing the latent sampling approaches studied in this work. Expected trends in reconstruction error versus KL regularization weight for prior sampling, posterior sampling, and LFM sampling.}
    \label{fig:latent_sampling}
\end{figure}

The behavior of each sampling method is directly influenced by the degree of latent space regularization applied during VAE training, which is controlled by the  KL regularization weight, $\lambda_{\mathrm{KL}}$, in Equation \eqref{eq:vae_loss}. Figure \ref{fig:latent_sampling} shows the expected trends in generative sampling errors for each approach as a function of $\lambda_{\mathrm{KL}}$.  Here, we expect posterior sampling to yield the lowest error across all $\lambda_{KL}$, since the encoder is trained to minimize reconstruction loss. Prior sampling is expected to fail at small $\lambda_{KL}$ due to mismatch between the learned posterior and the fixed prior and also at high $\lambda_{KL}$ due to underfitting the data, with an optimal reconstruction/regularization balance somewhere between. LFM sampling is expected to close the gap between prior and posterior sampling, recovering generative performance even when the VAE posterior diverges from the prior. Furthermore, LFM provides more robustness to particular choices of  $\lambda_{KL}$ while VAE requires more careful tuning.  As $\lambda_{KL}$ increases, all three sampling methods converge as the VAE posterior becomes increasingly aligned with the prior, and the benefit of a flexible latent model diminishes. We identify three key gaps to quantify these behaviors, as depicted in Figure \ref{fig:latent_sampling}:
\begin{itemize}[itemsep=1pt, parsep=0pt]
	\item \textbf{Prior-posterior gap}: reflects the generative failure of standard constrained VAEs under low KL regularization when the prior and posterior diverge.
	\item \textbf{Latent modeling gap}: captures the discrepancy between our LFM samples and the best-case posterior reconstructions, and is primarily due to approximation error in learning the latent distribution with flow matching.
	\item \textbf{Value-added gap}: measures the improvement of our approach over prior sampling, representing the net benefit of latent generative modeling versus a simpler, constrained VAE.
\end{itemize}

We note that LFM sampling incurs additional overhead that is not required for prior sampling with a constrained VAE for both training a flow matching model and using it to generate latent vectors (see Section \ref{sec:latent_flow_matching}). However, this overhead is typically negligible relative to the cost of pre-training the VAE given the low dimensionality of the latent space where generative modeling is occurring. For example, LFM training required less than two minutes for all problems considered in the following section while sampling required less than one second per 1000 samples using a single A100 GPU node. See Appendix \ref{sec:appendix_implementation} for more details on the LFM implementation and hyperparameters used for each example.

\section{Numerical Results}\label{sec:experiments}

The proposed latent generative modeling approach using c-LFM is demonstrated for random field reconstruction and random field inference problems. Each type of problem is first illustrated with simplified demonstrations followed by real-world applications, see Table \ref{tab:example_details} for a summary of the examples. More implementation details can be found in Appendix \ref{sec:appendix_implementation}, with hyperparameter settings for the VAE and LFM provided in Tables \ref{tab:nn_architecture_details} and \ref{tab:fm_hyperparams}, respectively. The code to reproduce all examples is available at \url{https://github.com/nasa/random_field_c-lfm}.

\begin{table}[htbp]
    \centering
    \caption{Summary of example problems considered.}
    \label{tab:example_details}
    \begin{tabular}{@{} l l l l c c c @{}}
        \toprule
        \multirow{2}{*}{\textbf{Example}} & 
        \multirow{2}{*}{\makecell[l]{\textbf{Problem} \\ \textbf{Type}}} & 
        \multicolumn{2}{c}{\textbf{Constraint}} & 
        \multirow{2}{*}{\makecell[c]{\textbf{Latent} \\ \textbf{Dim. ($\mathbf{d_z}$)}}} & 
         \multirow{2}{*}{\makecell[c]{\textbf{Num. Train} \\ \textbf{Data  ($\mathbf{N}$)} }} & 
        \multirow{2}{*}{\makecell[c]{\textbf{Num. } \\ \textbf{Sensors  ($\mathbf{m}$)} }} \\
        \cmidrule(lr){3-4}
        & & \textbf{Type} & \textbf{Form} & & & \\
        \midrule
        4.1.1 & Reconstruction & Statistical & Covariance/Correlation            & $4$  & $10^3$     & $1$--$5$  \\
        4.1.2 & Reconstruction & Physical    & Positivity            & $4$  & $10^4$    & $3$--$11$ \\
        4.2.1 & Inference      & Physical    & Poisson Equation           & $1$  & $10^1$--$10^3$ & $25$    \\
        4.2.2 & Inference      & Physical    & Poisson Equation           & $2$  & $10^1$--$10^3$ & $25$    \\
        4.3   & Reconstruction & Statistical & Coherence             & $32$ & $5 \times 10^3$     & $40$    \\
        4.4   & Inference      & Physical    & Linear Elasticity PDE & $16$ & $10^1$--$10^3$ & $90$    \\
        \bottomrule
    \end{tabular}
\end{table}

\FloatBarrier
\subsection{Demonstrations: Random Field Reconstruction From Sparse Data}\label{sec:reconstruction_demos}

\begin{figure}[htbp]
    \centering
    \begin{subfigure}{0.48 \textwidth}
        \includegraphics[width=\linewidth]{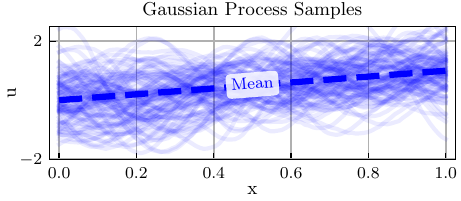}
        		\caption{} 
		\label{fig:sub1}
    \end{subfigure}
    \begin{subfigure}{0.48 \textwidth}
        \includegraphics[width=\linewidth]{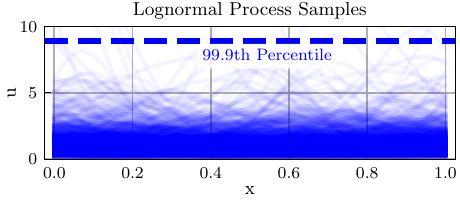}
        		\caption{} 
		\label{fig:sub2}
    \end{subfigure}
    \caption{Random fields considered for the reconstruction demonstrations in Section \ref{sec:reconstruction_demos}: a) 100 samples of a Gaussian process and b) 1000 samples of a lognormal process.}
    \label{fig:reconstruction_demo}
\end{figure}

We first demonstrate the proposed approach for reconstructing one-dimensional random fields (i.e., stochastic processes) using sparse observations. First, a Gaussian Process (GP) is considered to illustrate the utility of incorporating a statistical (covariance) constraint to supplement sparse data in Section \ref{sec:gp_recon}. Next, a Lognormal Process (LNP) is used to highlight the advantage of a latent generative modeling approach in a non-Gaussian, yet simple, setting in Section \ref{sec:lnp_recon}. Figure \ref{fig:reconstruction_demo} shows true samples and statistics for these two cases.

\subsubsection{Gaussian Process Reconstruction}\label{sec:gp_recon}

This demonstration considers a GP with squared exponential covariance function,
\begin{equation}
    \begin{aligned}
         & u(x) \sim \mathcal{GP}(\mu(x), \text{Cov}(x, x')), \;\; x\in[0, 1]                            \\
         & \text{Cov}(x, x') = \sigma^2 \text{exp}\left(-\frac{\|x-x'\|^2}{2 l^2}\right), \label{eq:gp_ex}
    \end{aligned}
\end{equation}
and linear mean function, $\mu(x)=x$, with variance, $\sigma^2=0.5$, and covariance length, $l=0.1$ (Figure \ref{fig:reconstruction_demo} a)). The goal is to learn to sample $u(x)$ from sparse ($m \leq 5$) equally-spaced sensors but with sufficient observations ($N=1000$). A statistical constraint, $R$, is used to impose the true covariance structure during training to supplement the sparse data,
\begin{equation}
	R(U(\{ \tilde{\mathbf{x}}^{(c)} \}, \omega)) = \hat{\mathbf{\Sigma}}^{gen}(\{ \tilde{\mathbf{x}}^{(c)} \}, \omega) - \mathbf{\Sigma}^{true}(\{ \tilde{\mathbf{x}}^{(c)} \}) = 0, \label{eq:gp_constraint}
\end{equation}
where $\hat{\mathbf{\Sigma}}^{gen}$ is the empirical covariance matrix computed with generated samples from c-LFM and $\mathbf{\Sigma}^{true}$ is the true covariance matrix formed using Equation \eqref{eq:gp_ex}. The constraint residual (and thus both covariance matrices) in Equation \eqref{eq:gp_constraint} are evaluated at $C=50$ randomly chosen collocation points, $\{ \tilde{\mathbf{x}}^{(c)} \}$, and weighted by $\lambda_r = 10^{-2}$ during training.

\begin{figure}[htbp]
    \centering
    \includegraphics[width=\textwidth]{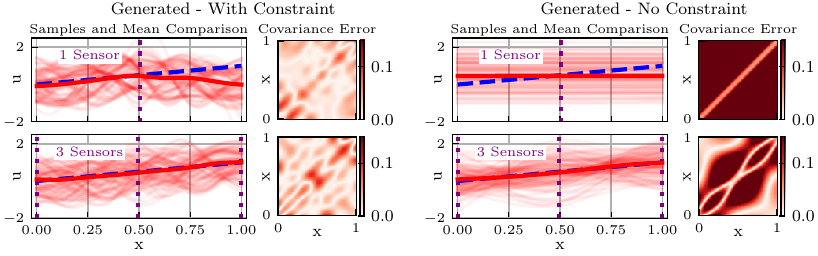}
    \caption{GP reconstruction from sparse sensors using c-LFM with covariance constraint (left) versus standard LFM with no constraint (right). The statistical constraint allows for accurate covariance recovery from sparse sensors.}
    \label{fig:gp_example_summary}
\end{figure}

Figure \ref{fig:gp_example_summary} shows the true GP samples and covariance and corresponding generated samples for $m=1, 3$, with and without a covariance constraint. c-LFM begins to generate visually similar samples and covariance to the true field with just a single sensor, while standard LFM generates spatially constant samples, as expected. Both approaches accurately capture the mean for $m=3$, while the no-constraint approach significantly underperforms the proposed approach in estimating covariance.

\begin{figure}[htbp]
    \centering
    \includegraphics[width=0.75\textwidth]{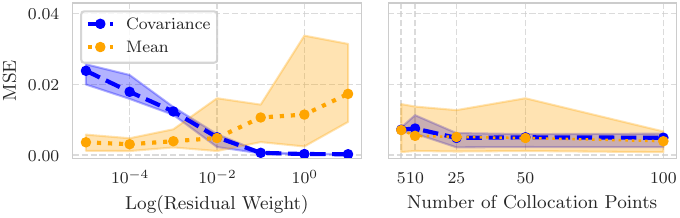}
    \caption{Errors of random field reconstruction when using a covariance constraint. Here the number of sensors is $m=3$ and (left) the number of collocation points $C=50$, (right) the residual weight is $\lambda_r=10^{-2}$.}
    \label{fig:gp_example_rw_and_nc}
\end{figure}

Figure \ref{fig:gp_example_rw_and_nc} shows the reconstruction accuracy as a function of residual weight, $\lambda_r$ (left), and number of collocation points, $C$ (right) for $m=3$ sensors. Here, the mean squared error (MSE) is calculated for the reconstructed covariance and mean functions using 1000 generated samples evaluated at 100 equally-spaced points. Each hyperparameter setting was evaluated across five random trials with the mean error indicated with lines and markers and the min/max errors with shading. The left plot demonstrates how the residual weight balances the enforcement of the statistical covariance constraint versus reconstructing the sparse data. As $\lambda_r$ increases, covariance MSE decreases due to the higher penalty placed upon the statistical residual, $R$, but the mean MSE increases due to less emphasis on reconstruction. Balanced errors can be seen for approximately $\lambda_r=10^{-2}$. Figure \ref{fig:gp_example_rw_and_nc} (right) illustrates that errors are relatively invariant to the number of collocation points, $C$, used to calculate the residual.  Using even a very small number of collocation points provides the necessary statistical regularization.

\begin{figure}[htbp]
    \centering
    \includegraphics[width=0.5\textwidth]{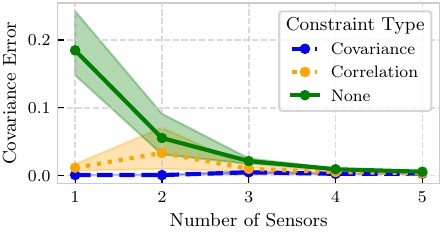}
    \caption{Comparison of constraint types on errors of random field reconstruction.  Here the number of collocation points is $C=50$ and the residual weight is $\lambda_r=10^{-2}$.}
    \label{fig:gp_example_num_sensors}
\end{figure}

The type of statistical constraint used for $R$ is compared in Figure \ref{fig:gp_example_num_sensors} as a function of the number of sensors, $m$, used in training.  The comparison is made between a residual based on covariance matching compared to a residual based on correlation matching.  The correlation constraint is the same as the covariance constraint without matching the magnitude (i.e. variance).  It is seen that both constraints provide benefit over the no-constraint case.  However, the correlation constraint, which lacks the magnitude component, must rely on more sensor measurements in order to estimate that portion of the signal.

\begin{figure}[htbp]
    \centering
    \includegraphics[width= 0.75 \textwidth]{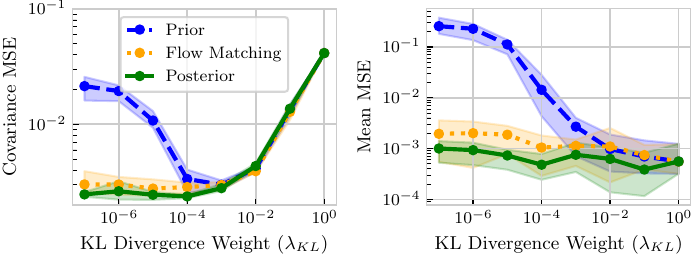}
    \caption{Performance of different sampling approaches on GP reconstruction accuracy as a function of KL Divergence weight in terms of (left) covariance MSE and (right) mean MSE. }
    \label{fig:gp_sampling_comparison}
\end{figure}

Finally, the GP reconstruction accuracy of c-LFM is compared to a constrained-VAE baseline ("Prior") and best-case VAE posterior sampling ("Posterior") in Figure \ref{fig:gp_sampling_comparison}. The left plot showing covariance MSE versus KL divergence weight for the three approaches reflects the expected trends depicted in Figure \ref{fig:latent_sampling}. Namely, posterior and c-LFM sampling accuracy improve with decreasing $\lambda_{KL}$ before plateauing, while prior sampling reaches a minimum error before increasing again for smaller $\lambda_{KL}$ values. This shows that the c-LFM approach is more robust to choice of $\lambda_{KL}$, provided it is small enough ($\lambda_{KL} \lesssim 10^{-3}$). However,  the value-added gap between c-LFM and the constrained-VAE baseline is shown to be relatively small, implying that a fixed, standard Gaussian VAE prior is expressive enough for accurate GP reconstruction for an appropriate choice of $\lambda_{KL}$. The next demonstration will show how this gap in performance widens in another simple setting but with non-Gaussian behavior.

\FloatBarrier
\subsubsection{Lognormal Process Reconstruction}\label{sec:lnp_recon}

This demonstration considers a LNP process,
\begin{equation}
         u(x)   \sim \text{exp}\left( \mathcal{GP}(0, \text{Cov}(x, x') ) \right), \;\; x\in[0, 1]                             \\
\end{equation}
with covariance function parameters: variance, $\sigma^2=0.5$, and correlation length, $l=0.2$. The goal here is to highlight the performance gap between c-LFM and a constrained VAE for reconstructing a simple non-Gaussian random field with heavy-tailed behavior (Figure \ref{fig:reconstruction_demo} b)).  We consider $3 \leq m \leq 11$ equally-spaced sensors and increase the number of observations to $N=10000$ to ensure sufficient data to capture the tail of the distribution. The statistical covariance constraint is removed and instead,  a positivity constraint, $F(u(x)) \geq 0$ for all $x$, is enforced during training using a rectified linear unit (ReLU) function with weight $\lambda_f=1.0$ to discourage samples with negative values. Accuracy is assessed in terms of covariance MSE and also the MSE in the estimated $99.9^{th}$ percentile to highlight effectiveness in capturing a heavy tailed distribution.  Each hyperparameter setting is again evaluated across five random seeds and the MSE is evaluated across $10000$ generated samples on a grid of $100$ equally-space points across the domain.

\begin{figure}[htbp]
    \centering
    \begin{subfigure}{0.95 \textwidth}
        \includegraphics[width=\linewidth]{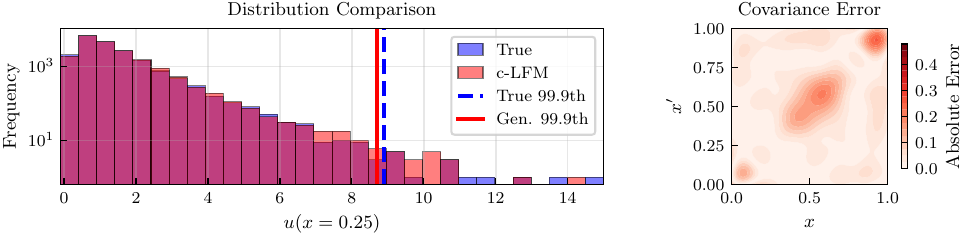}
        		\caption{} 
		\label{fig:sub1}
    \end{subfigure}
    \begin{subfigure}{0.95\textwidth}
        \includegraphics[width=\linewidth]{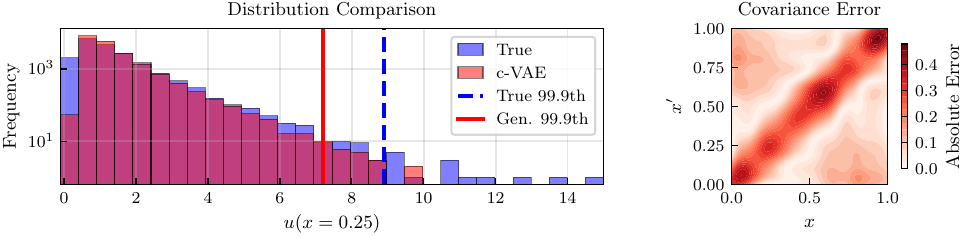}
        		\caption{} 
		\label{fig:sub2}
    \end{subfigure}
    \caption{Lognormal process reconstruction demonstration with $m=7$ sensors. A comparison of pointwise distributions of true vs. generated samples and absolute covariance errors for a) c-LFM and b) c-VAE. }
    \label{fig:lnp_demo}
\end{figure}

VAEs were trained using data from $3 \leq m \leq 11$ sensors along with the positivity constraint for a range of KL divergence weights, $\lambda_{KL} \in [10^{-6},  10^{-5}, ..., 10^{0}]$. Figure \ref{fig:lnp_demo} compares the best case solutions for $m=7$ sensors for a) c-LFM and b) the constrained VAE, which were obtained for $\lambda_{KL}=10^{-3}$ and $\lambda_{KL}=10^{-1}$, respectively. Here, the generated versus true pointwise distributions at $x=0.25$ (which is half way between sensors) along with $99.9^{th}$ percentiles are compared and the absolute error in covariance is shown for both methods. The constrained VAE struggles to capture the heavy-tailed behavior of the LNP and has significant percentile and covariance errors, while latent space sampling with c-LFM significantly improves the accuracy in this case.

\begin{figure}[htbp]
    \centering
    \begin{subfigure}{0.75 \textwidth}
        \includegraphics[width=\linewidth]{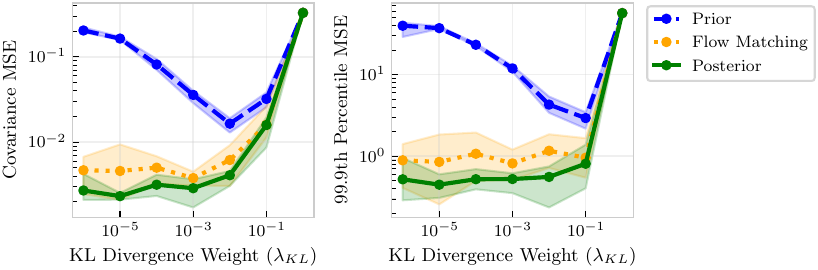}
        		\caption{} 
		\label{fig:sub1}
		    \vspace{8pt}
    \end{subfigure}
    \begin{subfigure}{0.75 \textwidth}
        \includegraphics[width=\linewidth]{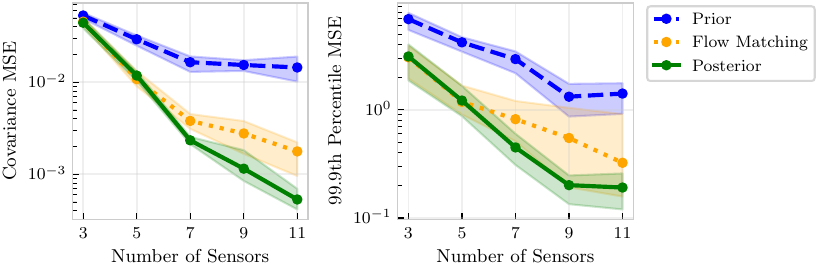}
        		\caption{} 
		\label{fig:sub2}
    \end{subfigure}
    \caption{Errors versus (a) KLD Weight for the case of $m=7$ sensors and (b) number of sensors for the best-case KLD Weight. }
    \label{fig:lnp_kld_sweep}
\end{figure}

Trends in LNP reconstruction errors as a function of KL divergence weight $\lambda_{KL}$ and number of sensors $m$ are shown in Figure \ref{fig:lnp_kld_sweep}, comparing the three sampling approaches described in \ref{sec:sampling_approaches}. The shortcomings of the constrained VAE are again highlighted by the consistently larger errors for prior sampling across all values of $\lambda_{KL}$ and $m$ considered. It appears that less regularization (i.e., lower $\lambda_{KL}$) is required to effectively capture the non-Gaussian LNP behavior. Furthermore, the performance of c-LFM is relatively robust to choice of $\lambda_{KL}$, with values across the range $[10^{-6}, 10^{-2}]$ shown to yield accurate results. While the resulting prior-posterior mismatch negatively affects the constrained VAE accuracy, c-LFM is able to reasonably approximate the more complex true posterior in these cases. Comparing the GP and LNP reconstruction performance, a significantly larger prior-posterior gap and value-added gap (c-LFM vs. constrained VAE) is observed for LNP, highlighting the benefits of the proposed approach for non-Gaussian, heavy-tailed settings.

\FloatBarrier
\subsection{Demonstrations: Random Field Inference from Indirect Data}\label{sec:inference_demos}

The ability of the proposed approach to perform random field inference is now demonstrated in a simple one-dimensional setting. A spatially-varying, random coefficient, $V(x, \omega)$, is estimated from indirect observations of the random field, $U(x, \omega)$, where samples of the two are related through the Poisson equation:
\begin{equation}
    \begin{aligned}
         & \frac{d}{dx}\left(v(x) \frac{du(x)}{dx}\right) = -sin(x), \;\; x \in [0, \pi], \label{eq:1dpoisson_ex} \\
         & v(x) = 1 + \epsilon x^2, \\
         & u(0) = 0, \; u(\pi) = 0.
    \end{aligned}
\end{equation}
Here, $\epsilon$ is a realization of a random variable that yields a stochastic Poisson problem. The goal is to learn to sample from $p(u, v)$ from varying numbers of observations ($10 \leq N \leq1000$) of $u(x)$ with sufficiently dense sensors ($m=25$). Synthetic training data is generated by sampling $\epsilon$, computing $v(x)$, and using numerical integration to solve for $u(x)$. The unknown random field, $V(x, \omega)$, is learned by incorporating Equation \ref{eq:1dpoisson_ex} into the loss via physical constraint, $F(u(x), v(x)) = 0$, using autodifferentiation. 

\begin{figure}[htbp]
    \centering
    \begin{subfigure}{0.49 \textwidth}
        \includegraphics[width=\linewidth]{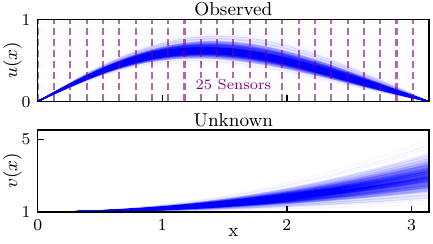}
        		\caption{} 
		\label{fig:sub1}
    \end{subfigure}
    \begin{subfigure}{0.49 \textwidth}
        \includegraphics[width=\linewidth]{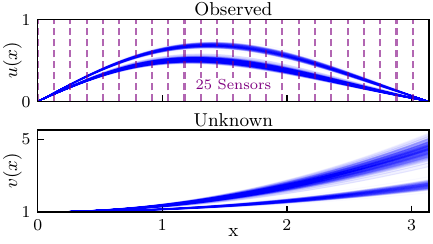}
        		\caption{} 
		\label{fig:sub2}
    \end{subfigure}
    \caption{Random fields considered for the inference demonstrations in Section \ref{sec:inference_demos}. Samples of an observed random field, $U(x, \omega)$, and unknown random coefficient, $V(x, \omega)$, that are related through the one-dimensional Poisson Equation for a) unimodal and b) bimodal cases.}
    \label{fig:inference_demo}
\end{figure}

Inference with c-LFM is demonstrated in two settings: 1) $V(x, \omega)$ is unimodal, illustrating inference from limited data ($N \leq 100$) in a simple setting (Section \ref{sec:unimodal_inference}), and 2) $V(x, \omega)$ is bimodal, highlighting the advantage of a more flexible, latent generative modeling approach versus using a constrained VAE directly (Section \ref{sec:bimodal_inference}). Figure \ref{fig:inference_demo} shows samples of $U(x, \omega)$ and $V(x, \omega)$ for each case.

\subsubsection{Unimodal Process Inference}\label{sec:unimodal_inference}

\begin{figure}[htbp]
    \centering
    \includegraphics[width=0.95 \textwidth]{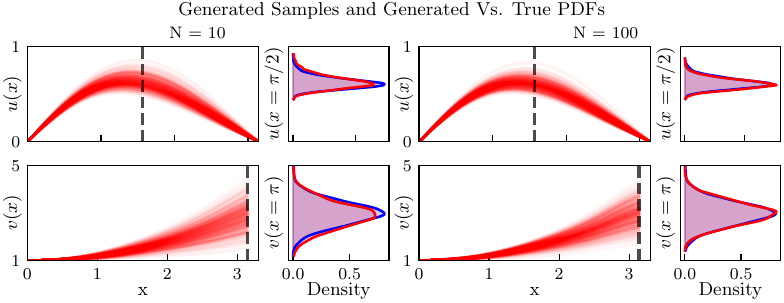}
    \caption{Demonstration of random field inference from indirect data with c-LFM for the unimodal case for $N=10$ and $N=100$ observations. Generated samples along with true (blue) versus generated (red) pointwise distributions. A physical constraint incorporating the Poisson equation allows for inference of $v(x)$ from limited observations of $u(x)$ only.}
    \label{fig:poisson1d_example_summary}
\end{figure}

In this case, randomness is introduced to the Poisson problem (Equation \eqref{eq:1dpoisson_ex}) via a Gaussian random variable, $\epsilon \sim \mathcal{N}(0.2, 0.05)$, yielding unimodal stochastic processes for both $U (x, \omega)$ and $V(x, \omega)$ (see Figure \ref{fig:inference_demo}  a)). Figure \ref{fig:poisson1d_example_summary} shows a demonstration of learning to sample $p(u, v)$ with c-LFM using only $N=10$ (left) and $N=100$ (right). Here, the objective function weights, $\lambda_f=10^{-3}$ and $\lambda_{KL}=10^{-6}$ were selected with manual hyperparameter tuning and $C=50$ collocation points were used to enforce the physical constraint, $F$. Sample-based empirical PDFs of $U(x=\pi/2)$ and $V(x=\pi)$ are shown to compare the point-wise distributions of the estimated versus true fields. Despite having no direct observations, $V(x, \omega)$ is recovered effectively using c-LFM with physical constraint, allowing reasonably accurate inference with only $N=10$ observations.

The sensitivity of c-LFM to variations in hyperparameter values was investigated. Specifically, the mean and variance MSE were tracked as a function of residual weight, $\lambda_f$, number of collocation points, $C$, and number of data points, $N$. The mean and variance functions of $U(x, \omega)$ and $V(x, \omega)$ were estimated with 1000 generated samples on 100 equally-spaced points. It was seen that there was little sensitivity to these parameters with two notable exceptions. First, performance degrades at large residual weights ($\lambda_f \geq 1.0$) and second, there was a consistent trend of more accurate results as the number of data points increases.

\begin{figure}[htbp]
    \centering
    \includegraphics[width= 0.75 \textwidth]{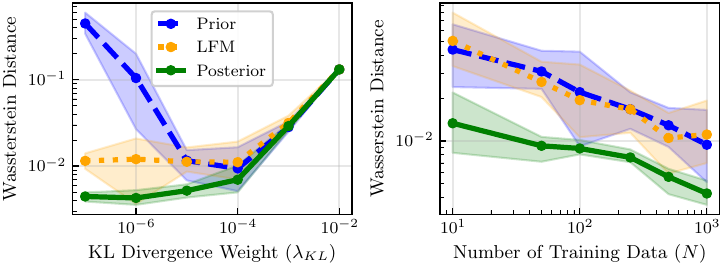}
    \caption{Performance of different sampling approaches on random field inference for the unimodal case. The Wasserstein distance between the generated and true random coefficient $V(x, \omega)$ is shown as a function of KL Divergence weight, $\lambda_{KL}$,  (left) and number of data, $N$(right).}
    \label{fig:poisson1d_unimodal_sampling_comparison}
\end{figure}

Finally, the inference accuracy for c-LFM is compared to a constrained-VAE baseline ("Prior") and best-case VAE posterior sampling ("Posterior") in Figure \ref{fig:poisson1d_unimodal_sampling_comparison}. Here, the average pointwise Wasserstein distance between the true and generated process, $V(x, \omega)$, is used to quantify inference accuracy and is viewed as a function of KL divergence weight (left) and number of observations (right). The expected trends are observed for each method: 1) LFM and posterior sampling accuracy improves before plateauing for decreasing $\lambda_{KL}$, 2) prior sampling fails for larger and smaller $\lambda_{KL}$ with an optimal balance achieved between, and 3) all methods gradually increasing in error for decreasing $N$. While the performance of c-LFM is shown to be more robust to the choice of $\lambda_{KL}$, the best-case inference accuracy for c-LFM and the constrained VAE baseline is comparable (negligible value-added gap), implying that a fixed VAE prior is sufficient in this 1D, unimodal setting. The next demonstration, however, shows a clear performance gap in a bimodal setting where c-LFM allows improved inference accuracy.

\FloatBarrier
\subsubsection{Bimodal Process Inference}\label{sec:bimodal_inference}


A mixture of Gaussians,  $\epsilon \sim w_1 \mathcal{N}(0.015, 0.035) + w_2 \mathcal{N}(0.15, 0.35)$ with $w_1=0.3$ and $w_2=0.7$ in Equation \eqref{eq:1dpoisson_ex}, is now introduced to yield bimodal behavior in $U (x, \omega)$ and $V(x, \omega)$  (see Figure \ref{fig:inference_demo}  b)). In this setting, we will highlight the utility of a latent-space sampling approach offered by c-LFM versus the constrained VAE baseline. To facilitate a fair comparison, a grid-based hyperparameter optimization is performed over several orders of magnitudes of objective function weights, $\lambda_{KL} \in [10^{-7}, ..., 10^{-1}]$ and $\lambda_{f} \in [10^{-6}, ..., 10^{0}]$, and the best-case performance for each method is identified. The Wasserstein distance between generated and test data samples of $V(x, \omega)$ is again used to assess inference accuracy.

\begin{figure}[htbp]
    \centering
    \includegraphics[width=0.95 \textwidth]{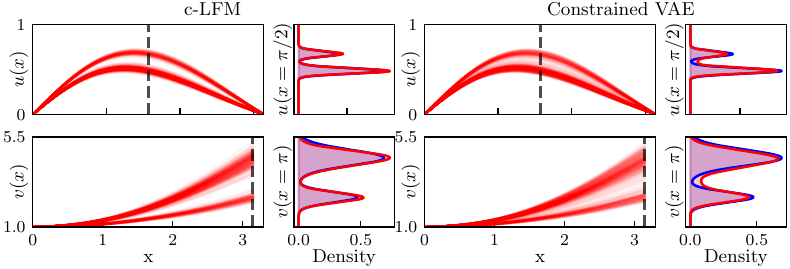}
    \caption{Demonstration of random field inference from indirect data for the bimodal case, comparing c-LFM (left) to prior sampling via a constrained VAE (right). Generated samples are shown along with a comparison of true (blue) versus generated (red) pointwise distributions. }
    \label{fig:poisson1d_bimodal_example_summary}
\end{figure}

Figure \ref{fig:poisson1d_bimodal_example_summary} compares generated samples of $U(x, \omega)$ and $V(x, \omega)$ for the case of $N=1000$ observations with c-LFM (left) versus a constrained VAE (right) along with comparisons of pointwise distributions of generated vs. true samples for each method. Here, the optimal weights were $\lambda_{KL} = 10^{-6}$ and $\lambda_{f} = 10^{-1}$ for c-LFM and $\lambda_{KL} = 10^{-4}$ and $\lambda_{f} = 10^{-1}$ for the constrained VAE. This result shows qualitatively that c-LFM is able to more cleanly identify the two distinct modes of $V(x, \omega)$ relative to the constrained VAE, which generates many samples lying in the near-zero-probability region of the true distribution.

\begin{figure}[htbp]
    \centering
    \includegraphics[width= 0.75 \textwidth]{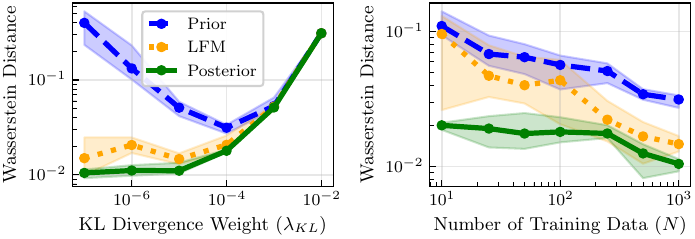}
    \caption{Performance of different sampling approaches on random field inference for the multimodal case. The Wassterstein distance between the generated and true random coefficient $V(x, \omega)$ is shown as a function of KL Divergence weight, $\lambda_{KL}$,  (left) and number of data, $N$, (right).}
    \label{fig:poisson1d_multimodal_sampling_comparison}
\end{figure}

Inference errors for c-LFM and constrained VAE as a function of KL divergence weight $\lambda_{KL}$ and number of observations $N$ are shown in Figure \ref{fig:poisson1d_multimodal_sampling_comparison}, compared to the best-case posterior sampling. Compared to the previous unimodal case, there is a significantly larger prior-posterior gap and value-added gap provided by c-LFM observed in the plot of Wasserstein distance versus $\lambda_{KL}$ when considering bimodal random fields. Again, the result highlights the robustness of c-LFM to choice of $\lambda_{KL}$ relative to that of the constrained VAE. The plot of error versus number of observations indicates that c-LFM outperforms the constrained VAE across the range of $N$ considered. However, the performance gap closes for lower numbers of observations ($N \leq 100$).

\begin{figure}[htbp]
    \centering
    \includegraphics[width= 0.98 \textwidth]{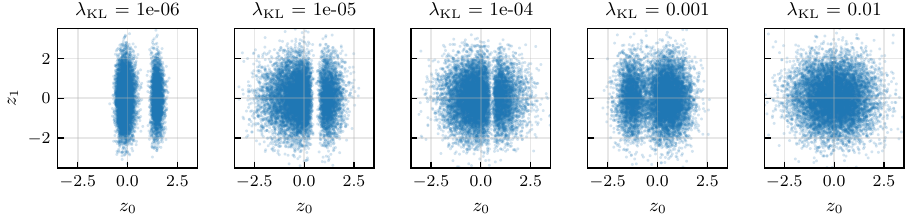}
    \caption{Visualization of the learned latent space for varying $\lambda_{KL}$ in the multimodal random field inference demonstration. }
    \label{fig:poisson1d_multimodal_latent_space_comparison}
\end{figure}

Figure \ref{fig:poisson1d_multimodal_latent_space_comparison} provides a visualization of the learned posterior distribution of latent variables for varying KL divergence weights. For $\lambda_{KL} = 10^{-2}$ (higher regularization), the latent space appears similar to the VAE prior (standard Gaussian), but as $\lambda_{KL}$ decreases, two distinct modes form. This illustrates how a smaller KL divergence weight can provide greater latent space flexibility for reducing data reconstruction error and constraint residuals during VAE training. In this case, the latent space tends toward a bimodal distribution in order to more accurately capture the bimodal random field being inferred. Sampling the latent prior (i.e., using the constrained VAE baseline) yields higher errors due to the mismatch with the bimodal posterior, in part due to producing samples in the low (or zero) probability region between modes (as observed in Figure \ref{fig:poisson1d_bimodal_example_summary}). On the other hand, latent generative modeling with c-LFM allows the posterior to be accurately sampled even when it is more complex due to less regularization.

\FloatBarrier
\subsection{Application: Wind Velocity Field Estimation}\label{sec:windEstimation}

\begin{figure}[htbp]
	\centering
	\begin{subfigure}{0.28 \textwidth}
		\includegraphics[width=\linewidth]{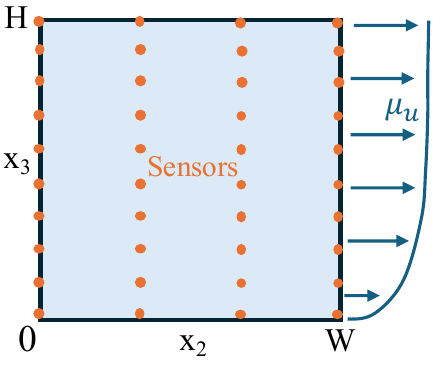}
		\caption{} 
		\label{fig:wind_domain}
	\end{subfigure}
	\hspace{0.1\textwidth}
	\begin{subfigure}{0.27 \textwidth}
    		\includegraphics[width=\linewidth]{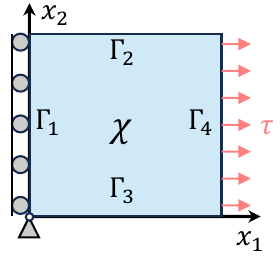}
				\caption{} 
   		 \label{fig:materials_domain}
	\end{subfigure}
	\caption{Domain diagrams for the applications considered: (a) wind velocity estimation (b) material property characterization. }
	\label{fig:domain_diagrams}
\end{figure}

A real-world example of random field reconstruction is presented to demonstrate c-LFM in a higher-dimensional setting with a more complex statistical constraint. The goal is to generate realistic samples of a wind velocity field, $\mathbf{U}(\mathbf{x}, \omega)$, from sparse spatio-temporal observations. Synthetic measurement data is generated via Monte Carlo simulation following a wind engineering formulation \citep{CARASSALE2006323} that decomposes a random wind field as
\begin{equation}
	\mathbf{U}(\mathbf{x}, \omega) = \boldsymbol{\mu}_u(\mathbf{x}) + \mathbf{W}(\mathbf{x}, \omega), \label{eq:wind_field}
\end{equation}
where $\mathbf{x} = [x_1, x_2, x_3, t]$, $\mathbf{W}(\mathbf{x}, \omega)$ is a zero-mean, stationary GP, and $\boldsymbol{\mu}_u$ is a deterministic mean function describing a logarithmic profile versus altitude, 
\begin{equation}
	\boldsymbol{\mu}_u(\mathbf{x}) = 2.5 u_* \text{ln}\left(\frac{x_3}{z_0} \right). \label{eq:wind_mean}
\end{equation}
Here, $u_* = 1.8$ m/s is the shear velocity and $z_0 =0.015$ m is the surface roughness. The spatial and temporal correlation of $\mathbf{W}$ (and hence $\mathbf{U}$) is prescribed by the \textit{coherence function}:
\begin{equation}
	\text{Coh}(\mathbf{x}, \mathbf{x}', n) = \text{exp}\left( - \frac{n \| \mathbf{c}^T (\mathbf{x} -  \mathbf{x}' )\|}{\| \mu_u(\mathbf{x}) - \mu_u(\mathbf{x}') \|} \right), \label{eq:coherence_function}
\end{equation}
which generalizes spatial correlation to include frequency, $n$. Decay coefficients, $c = [3, 3, 0.5]$, control coherence decrease in each spatial direction. Samples of $\mathbf{U}$ for training and test data were generated using the spectral representation method \citep{shinozuka1991simulation, DEODATIS2025110522} implemented in \texttt{UQpy} \citep{olivier2020uqpy}.

\begin{figure}[htbp]
	\centering
	\begin{subfigure}{0.7 \textwidth}
		\includegraphics[width=\linewidth]{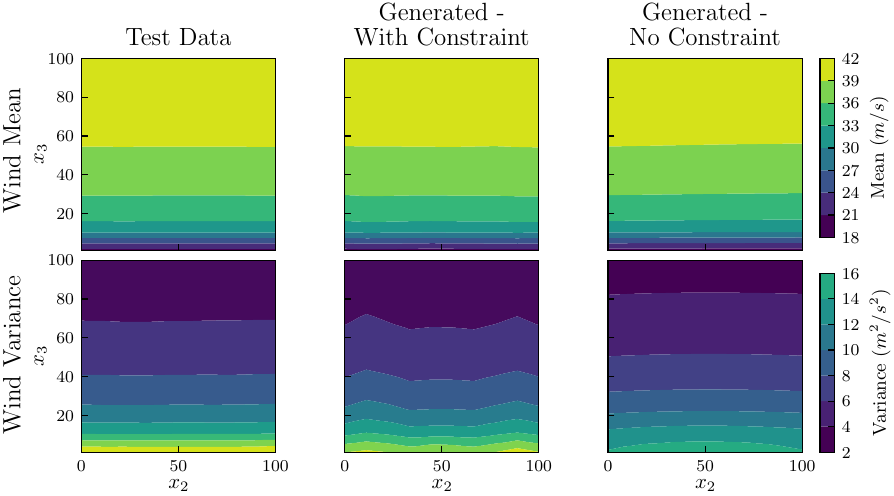}
		\caption{} 
		\label{fig:windstats}
	\end{subfigure}
	\begin{subfigure}{0.93\textwidth}
		\includegraphics[width=\linewidth]{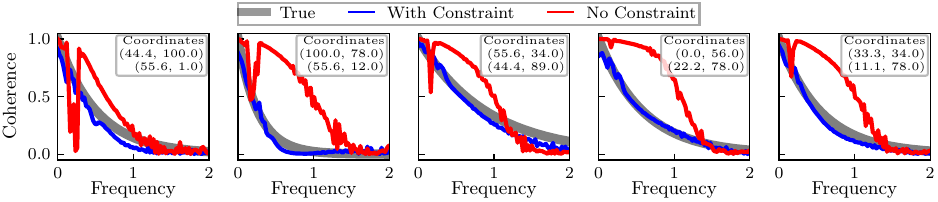}
		\caption{} 
		\label{fig:windcoherence}
	\end{subfigure}
	\caption{Wind velocity estimation (a) diagram along with comparisons of (b) mean and variance wind fields  and (c)  coherence  for the true wind test data versus generated samples (with and without coherence constraint). }
	\label{fig:windExample}
\end{figure}

This example focuses on reconstructing the $U_1$ velocity component over time in a two-dimensional plane, $[x_2, x_3, t] \times [0, 100]m \times [1, 100]m \times [0, 42.5]s$; see Figure \ref{fig:wind_domain}. The training data comprised $N=5000$ velocity samples at $256$ time steps and four equally-spaced vertical sensor arrays with ten sensors each, mimicking real-world wind profilers \citep{coulter2020sonic, shah2025generativemodelingmicroweatherwind}. c-LFM was applied with Equation \eqref{eq:coherence_function} used as a statistical constraint for training, where the coherence residual was estimated based on the \texttt{pytorch} fast Fourier transform. Importantly, since the coherence is a normalized metric, c-LFM must balance learning the wind velocity mean and variance from the sparse data and spatial/temporal correlation from the statistical constraint. See Appendex \ref{sec:wind_appendix} for more details on training data generation and the coherence constraint implementation.

The proposed approach was trained with objective function weights, $\lambda_r =10^{-2}$ and $\lambda_{KL} =10^{-7}$, and $C=16$ pairs of collocation points used to estimate the coherence residual. Figure \ref{fig:windstats} shows the mean and variance fields estimated from 1000 samples and compares generated samples with and without a coherence constraint to true samples from a test dataset. Generated samples for both cases are similar, accurately recovering the mean but slightly underestimating the variance near $x_3=0$ where test data variance increases rapidly. Figure \ref{fig:windcoherence} compares true wind coherence to empirical estimates from generated samples at five spatial coordinate pairs. Standard LFM struggles while c-LFM accurately recovers the coherence from the statistical constraint.

\begin{figure}[htbp]
	\centering
	\includegraphics[width=\textwidth]{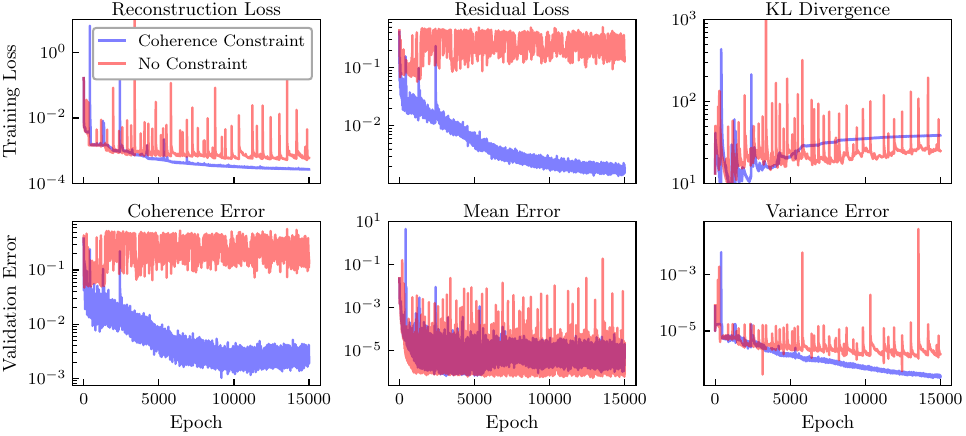}
	\caption{VAE convergence for the wind velocity estimation application. The loss function terms (top) and validation metrics (bottom) versus epoch both with and without the coherence constraint.}
	\label{fig:windConvergence}
\end{figure}

Figure \ref{fig:windConvergence} compares the convergence of LFM both with and without the coherence constraint for both the training loss terms (top) and validation data metrics (bottom). The addition of the coherence constraint appears to add regularization and stability to the training process, as evidenced by the periodic spikes in the loss function for the conventional LFM that are not present with the constraint. Both approaches achieve similar error in the mean estimates on the validation set, but c-LFM significantly reduces the coherence error while also improving the accuracy of variance estimates.

Table \ref{tab:wind_vae_vs_lfm} compares the errors in estimated wind velocity statistics from sampling with LFM versus sampling the pre-trained VAE directly both with and without the coherence constraint. The statistics used for the error calculations were estimated from 1000 samples for both the generated and test data. LFM with coherence constraint (c-LFM) shows superior coherence and variance accuracy with mean MSE only slightly higher than the unconstrained LFM approach. It can be seen that both the constrained and unconstrained VAE have consistently higher errors across the three metrics relative to their LFM counterparts.

\begin{table}[htbp]
    \centering
    \caption{Errors in wind statistics for sampling with LFM versus the pre-trained VAE ("Prior" sampling) both with and without the coherence constraint.}
    \label{tab:wind_vae_vs_lfm}
    \begin{tabular}{lcccc}
    \toprule
  &   \textbf{Sampling} & \textbf{Coherence MSE} & \textbf{Mean MSE} & \textbf{Variance MSE} \\
    \midrule
   With Constraint & LFM & \textbf{7.20e-03} & 4.46e-02 & \textbf{9.82e-01} \\ 
    & VAE & 1.11e-02 & 6.49e-02 & 1.13e+00 \\
 \midrule
 No Constraint & LFM & 1.06e-01 & \textbf{4.33e-02} & 7.13e+00 \\
 &VAE & 1.18e-01 & 7.78e-02 & 2.14e+01 \\
\bottomrule
    \end{tabular}
    \end{table}

\begin{figure}[htbp]
	\centering
	\includegraphics[width=\textwidth]{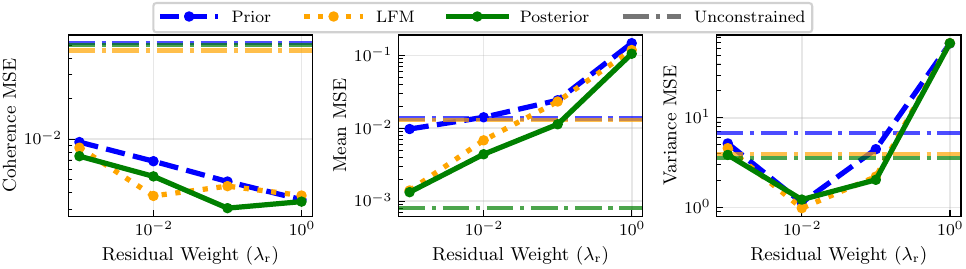}
	\caption{Performance of different sampling approaches for wind velocity estimation in terms of coherence (left), mean (middle), and variance (right) MSE. Errors are shown as a function of the coherence constraint residual weight with the unconstrained version of each approach ($\lambda_r=0$) is shown as a baseline.}
	\label{fig:windErrorsVsResWeight}
\end{figure}

Finally, an ablation study was performed by training models across a range of objective function weights $\lambda_{r} \in \{ 0.0, 10^{-3}, 10^{-2}, 10^{-1}, 10^{0} \}$, and KL divergence weight, $\lambda_{KL} \in \{10^{-8}, 10^{-7}, 10^{-6}, 10^{-5}, 10^{-4} \}$, where $\lambda_{r} = 0.0$ represents the no coherence constraint case. Figure \ref{fig:windErrorsVsResWeight} shows the resulting errors as a function of the residual weight ($\lambda_{r}$) for c-LFM versus both prior and posterior sampling with a constrained VAE. The unconstrained version of each sampling approach is also shown as a baseline. Here, the reported errors are for the best case (minimum error) value of $\lambda_{KL}$, averaged across three random seeds.  It can be seen that the coherence MSE decreases while the mean MSE increases with increasing residual weight and more emphasis on enforcing the coherence constraint, as expected. Interestingly, the variance MSE is minimized for $\lambda_{r} = 10^{-2}$, with increasing errors for smaller and larger values, indicating that some care is needed in adjusting objective function weights for an optimal balance of errors. Naturally, the coherence errors for all constrained approaches are significantly lower than their unconstrained counterparts, while the relative performance for mean and variance MSE is dependent on choice of $\lambda_{r}$. Finally, c-LFM generally outperforms the constrained VAE ("Prior" sampling) but the difference is not substantial, potentially due to the Gaussian nature of the randomness that is modeled effectively by the VAE's Gaussian latent space.

\FloatBarrier
\subsection{Application: Material Property Characterization}\label{sec:material_property}


A real-world engineering inference problem is presented in which only indirect observations are available. The goal is to estimate a random, spatially-varying elastic modulus field, $V(\mathbf{x}, \omega)$, using measurements of the mechanical deformation, $\mathbf{U}(\mathbf{x}, \omega)$, induced by a deterministic tensile force. We consider the two-dimensional domain illustrated in Figure \ref{fig:materials_domain} where $\mathbf{x} \in [0, 1] \times [0, 1]$ and $\mathbf{u} \in \mathbb{R}^2$.  The governing PDE that relates samples of $\mathbf{U}$ and $V$ is
\begin{equation}
	\mathcal{N}_x(\mathbf{u}, v) = \nabla \cdot \boldsymbol{\sigma}(\mathbf{u}, v) = \mathbf{0}, \label{eq:materials_pdf}
\end{equation}
where $\boldsymbol{\sigma}$ is the second-order stress tensor. Assuming a two-dimensional plane-stress formulation, isotropic elasticity, and small strain theory  \citep{reddy_2013}, the stress tensor is 
\begin{equation}
    \boldsymbol{\sigma} = \frac{2V}{1+\upsilon} \bigg[\nabla \mathbf{u} + \nabla \mathbf{u}^T + \frac{2 \upsilon}{1-\upsilon} \mathrm{Tr}(\nabla \mathbf{u})\mathbf{I}\bigg]
\end{equation}
where $\upsilon=0.3$ is the Poisson ratio and $\mathbf{I}$ is the identity tensor. The boundary conditions are $u_1 = 0$ for $\mathbf{x} \in \Gamma_1$, $u_2 = 0$ for $\mathbf{x} = (0, 0)$ and
\begin{equation}
	\boldsymbol{\sigma}(\mathbf{u}, V) \cdot \mathbf{n} = \boldsymbol{\tau}
\end{equation}
where $\mathbf{n}$ is the boundary outward normal, $\boldsymbol{\tau} = [0, 0]$ on $\Gamma_2$ and $\Gamma_3$ and $\boldsymbol{\tau} = [1.5, 0]$ on $\Gamma_4$. We note that while an isotropic elasticity setting was considered here for simplicity, future extensions of c-LFM could attempt to incorporate PINN approaches for handling anisotropy \citep{ZHAO2025107665} and plasticity  \citep{HAGHIGHAT2021113741}.

In this application, the random field $V(\mathbf{x}, \omega)$ represents the aleatory uncertainty associated with a family of test articles with spatial and test-to-test variability. Relatively dense measurements are feasible in real-world applications through the use of digital image correlation \citep{sutton2009image}, but the total number of test articles is often $O(10) - O(100)$. The ground truth material property was modeled as the lognormal random field, implemented by transforming a zero-mean Gaussian process using an exponential function,
\begin{align}\label{eq:matlTransform}
    V(\mathbf{x}, \omega) &= \alpha + \beta \exp(g(\mathbf{x}, \omega)), \\ 
    g(\mathbf{x}, \omega) &\sim \mathcal{GP}\bigg(0, \exp (-\frac{\|\mathbf{x}-\mathbf{x}'\|^2}{2l^2})\bigg),
\end{align}
with $\alpha=1.0$, $\beta=0.1$, and correlation length $l=1.0$ \citep{warner2020inverseestimationelasticmodulus}. Samples were generated by computing a truncated Karhunen-Loeve (KL) expansion \citep{fukunaga1970application} for $g(\mathbf{x}, \omega)$ and then applying the transformation in Equation \ref{eq:matlTransform}. A five-term KL expansion was used, retaining $\approx99\%$ of the variance of $g(\mathbf{x}, \omega)$.

Synthetic training data was obtained by evaluating each generated sample of $V(\mathbf{x}, \omega)$ using a forward solve of a finite element model implemented in the \texttt{FEniCS} Python library to compute corresponding samples of $\mathbf{U}(\mathbf{x}, \omega)$. Displacement measurement data were extracted on a $(10\times10)$ uniform grid spanning the domain with left column ($\mathbf{x}\in \Gamma_1$) removed, resulting in $m=90$ sensors. To study the effect of limited data at train time, we vary the number of training data, $N=10, 50, 100, 250, 500, 1000$ and use $10000$ samples for testing. 

Inference was facilitated through the use of a physical constraint that was a weighted combination of the PDE and applied boundary condition residuals: 
\begin{equation}
	F(\mathbf{u}, v) = \lambda_{N} \mathcal{N}_x(\mathbf{u}, v) + \lambda_{B_d}\mathcal{B}_d(\mathbf{u}) + \lambda_{B_n}\mathcal{B}_n(\mathbf{u}, v) \label{eq:mat_constraint}
\end{equation}	
where  $\lambda_{N}$, $\lambda_{B_d}$, and $\lambda_{B_n}$ are weights controlling the relative importance of the PDE, Dirichlet boundary condition, and Neumann boundary condition residuals while minimizing the VAE loss (Equation \eqref{eq:vae_loss}). In this case, the total physics residual weight was fixed as $\lambda_f=1.0$. It is well known that appropriately balancing various loss terms represents a fundamental challenge for training physics-constrained neural networks \citep{WANG2022110768}. In this work, we performed a simple, but expensive, grid search for each combination of physics residual weights, $\lambda_{N} \in \{10^{-5}, 10^{-4}, 10^{-3}, 10^{-2} \}$, $\lambda_{B_d} \in \{10^{-1}, 10^{0} \}$, $\lambda_{B_n} \in \{10^{-4}, 10^{-3}, 10^{-2} \}$, and KL divergence weight, $\lambda_{KL} \in \{10^{-8}, 10^{-7}, ..., 10^{-3} \}$ for a single random seed. From here, the best-case hyperparameters were selected as those that resulted in the minimum Wasserstein distance between the generated and true elastic modulus, $V(x, \omega)$. The best case hyperparameters were then evaluated across five random trials. It is expected that more advanced adaptive approaches for selecting optimal weights \citep{WANG2024116813,10.1016/j.neucom.2022.05.015} would improve the performance, but this is left for future work. 

\begin{figure}[htbp]
	\centering
	\begin{subfigure}{0.98\textwidth}
		\includegraphics[width=\linewidth]{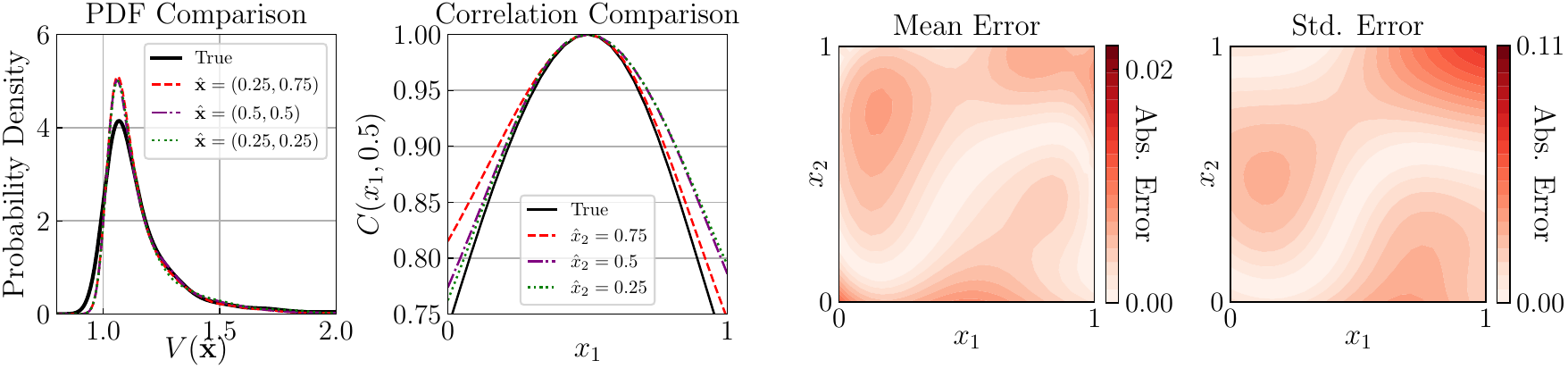}
		\caption{} 
		\label{fig:lfm_material_results}
	\end{subfigure}
	\begin{subfigure}{0.98\textwidth}
		\includegraphics[width=\linewidth]{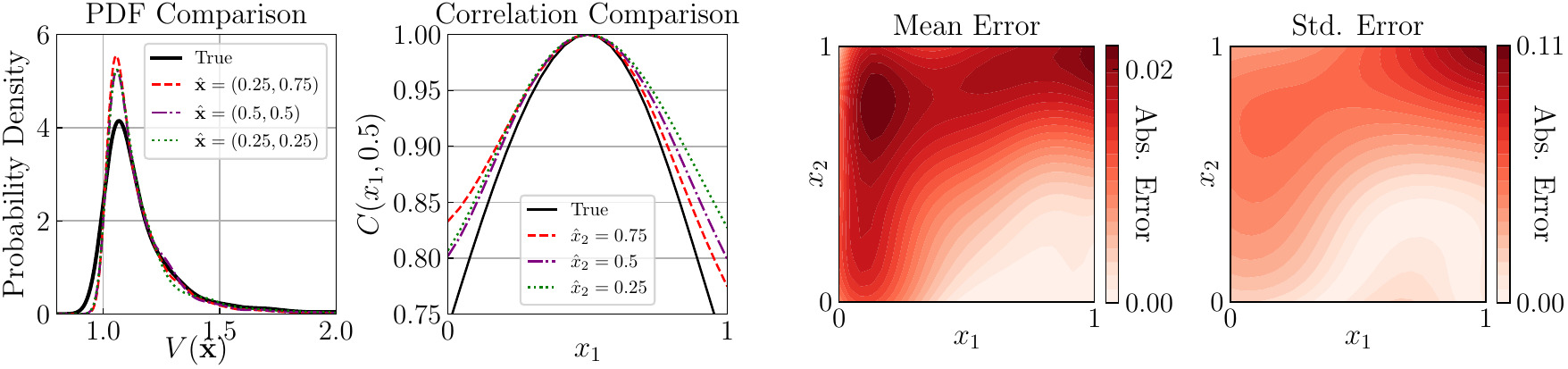}
		\caption{} 
		\label{fig:vae_material_results}
	\end{subfigure}
	\caption{A comparison of performance for material property inference for a) c-LFM versus b) c-VAE. From left to right, inference accuracy is view in terms of 1) generated PDFs at three distinct spatial coordinates, 2) the correlation with respect to $V(0.5, \hat{x_2})$ along horizontal slices, 3) mean field absolute error, and 4) standard deviations field absolute error. }
	\label{fig:materialResultsComparison}
\end{figure}

Results for the best performing models from the grid search for the case of $N=1000$ are compared in Figure \ref{fig:materialResultsComparison} for a) c-LFM ($\lambda_{N} = 10^{-3}, \lambda_{B_d} = 10^{-1}, \lambda_{B_n} = 10^{-3}, \lambda_{KL} = 10^{-6}$) and b) c-VAE ($\lambda_{N} = 10^{-3}, \lambda_{B_d} = 10^{0}, \lambda_{B_n} = 10^{-3}, \lambda_{KL} = 10^{-5}$). Here, the accuracy is viewed (from left to right), in terms of the estimated probability density functions (PDFs) at three distinct locations in the domain, the estimated spatial correlation across three horizontal slices  in the domain (Correlation$(x_1, 0.5)$), and the absolute errors in mean and standard deviation throughout the domain. It can be seen that c-LFM largely outperforms c-VAE for material property inference, with significantly better accuracy observed in the estimated mean and standard deviation functions.

\begin{figure}[htbp]
	\centering
	\begin{subfigure}{0.45\textwidth}
		\includegraphics[width=\linewidth]{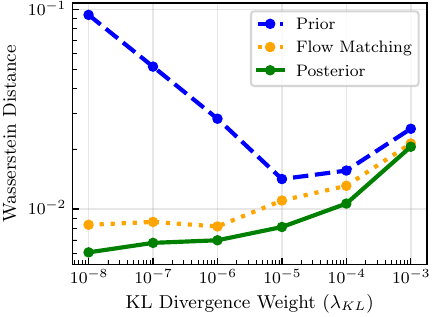}
		\caption{} 
		\label{fig:materials_error_vs_kld_weight}
	\end{subfigure}
	\begin{subfigure}{0.45\textwidth}
		\includegraphics[width=\linewidth]{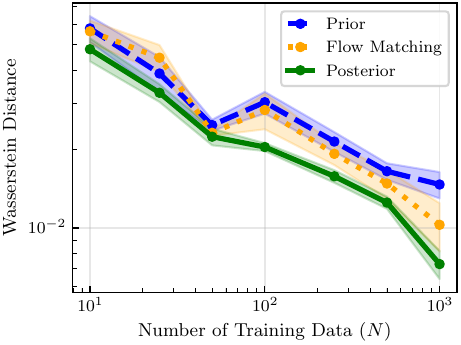}
		\caption{} 
		\label{fig:materials_error_vs_num_data}
	\end{subfigure}
	\caption{Performance of different latent sampling approaches for material property inference. The Wasserstein distance between the generated and true random elastic modulus, $V(\mathbf{x}, \omega)$, is shown as a function of (a) KL divergence weight for the bes-case physics residual weights for $N=1000$, and (b) number of training data for the best-case KL divergence and physics residual weights.}
	\label{fig:materialsSamplingComparison}
\end{figure}

Figure \ref{fig:materials_error_vs_kld_weight} compares the performance of the three latent sampling approaches (prior, flow matching, and posterior) for material property inference with $N=1000$ observations. Here, the Wasserstein distance between the true and generated elastic modulus field is viewed as a function of KL divergence weight. As expected, prior sampling (i.e., c-VAE) is the least accurate across the range of $\lambda_{KL}$ tested and also most sensitive to particular choices of the weighting. Compared to the simpler demonstration problems, there is a larger latent modeling gap between the flow matching and posterior sampling performance, indicating the increased complexity of the latent space in this application. 

Finally, Figure \ref{fig:materials_error_vs_num_data} shows the Wasserstein distance versus number of training data for each latent sampling approach using best-case hyperparameters. Inference performance gradually deteriorates with decreasing data for all three methods, while this deterioration is more rapid in the lowest data regimes tested ($10 \leq N \leq 50$).  It can be seen that c-LFM outperforms c-VAE for $N \geq 100$ while the two methods have comparable accuracy for $N \leq 50$. The crossover point in performance corresponds to where the amount of training data is similar to the latent dimension ($d_z = 16$) for this application, indicating that c-VAE may benefit from added latent space regularization to avoid overfitting in very low training data regimes. Strategies for maintaining inference accuracy for $N \leq 100$ for real-world applications will be the subject of future work.

\section{Conclusion}\label{sec:conclusion}

This work introduced a novel framework for latent generative modeling of random fields from limited training data, motivated by the unique challenges of science and engineering applications. The proposed approach addresses a critical gap in existing methods by enabling estimation of unknown distributions over functions (random fields) that capture aleatory uncertainty, even when only sparse or indirect data is available \textit{at train time}. Key features of the framework include the ability to produce continuous function samples in space and/or time and the generality to handle both reconstruction and inference problems using statistical or physical domain knowledge constraints.

The framework combines a constraint-aware VAE with a DeepONet function decoder for learning compact, physics- or statistics-informed latent representations, followed by generative modeling in the latent space. In this study, flow matching was used to sample the constraint-aware latent space and was termed c-LFM. A central contribution of this work is demonstrating why a latent-space approach is uniquely suited for constrained generative modeling in data-limited regimes. To the authors' knowledge, no existing multi-step generative method (e.g., diffusion or flow matching) is applicable when training data is sparse or indirect, while decoupling the learning of constraints from generative modeling with a latent space approach allows these powerful methods to be deployed. Moreover, through systematic comparisons with constrained VAE baselines, c-LFM was shown to achieve superior performance and was also significantly more robust to the choice of KL divergence regularization weight. Simple demonstrations were used to highlight cases where constrained VAEs fail to capture heavy-tailed or bimodal behavior in unknown random fields while c-LFM succeeds.

Efficacy was further demonstrated on two challenging applications. A wind velocity field estimation problem highlighted the importance of incorporating statistical constraints for capturing realistic spatial and temporal coherence from sparse sensors. The coherence constraint not only improved the realism of generated samples but also appeared to regularize the problem and improve VAE training convergence. A material property characterization problem demonstrated accurate inference of a non-Gaussian random field from indirect observations by incorporating physical (partial differential equation-based) constraints. Here, the proposed latent-space approach was again shown to achieve significantly better inference accuracy relative to using a constrained VAE directly.

The main limitation of the proposed work is that it did not directly address measurement noise in training data, another common challenge in science and engineering problems. Future work will extend the approach to settings with imperfect physical or statistical knowledge, manifesting as a mismatch between training data and imposed constraints, with noisy data being one such example. Additionally, the real-world applications will be studied further, where performance improvements are expected through more advanced neural network architectures within the VAE and flow matching models, adaptive training strategies to better balance loss terms, and more extensive hyperparameter tuning.


\FloatBarrier

\bibliographystyle{cas-model2-names}
\bibliography{lgm_cmame_preprint}

@article{Kingma2013AutoEncodingVB,
  title={Auto-Encoding Variational {Bayes}},
  author={Diederik P. Kingma and Max Welling},
  journal={CoRR},
  year={2013},
  volume={abs/1312.6114},
  url={https://api.semanticscholar.org/CorpusID:216078090}
}

@article{saxena2021generative,
  title={Generative adversarial networks ({GAN}s) challenges, solutions, and future directions},
  author={Saxena, Divya and Cao, Jiannong},
  journal={ACM Computing Surveys (CSUR)},
  volume={54},
  number={3},
  pages={1--42},
  year={2021},
  publisher={ACM New York, NY, USA}
}

@article{ho2020denoising,
  title={Denoising diffusion probabilistic models},
  author={Ho, Jonathan and Jain, Ajay and Abbeel, Pieter},
  journal={Advances in neural information processing systems},
  volume={33},
  pages={6840--6851},
  year={2020}
}

@inproceedings{nichol2021improved,
  title={Improved denoising diffusion probabilistic models},
  author={Nichol, Alexander Quinn and Dhariwal, Prafulla},
  booktitle={International conference on machine learning},
  pages={8162--8171},
  year={2021},
  organization={PMLR}
}

@article{lipman2022flow,
  title={Flow matching for generative modeling},
  author={Lipman, Yaron and Chen, Ricky TQ and Ben-Hamu, Heli and Nickel, Maximilian and Le, Matt},
  journal={arXiv preprint arXiv:2210.02747},
  year={2022}
}

@article{tong2023improving,
  title={Improving and generalizing flow-based generative models with minibatch optimal transport},
  author={Tong, Alexander and Malkin, Nikolay and Huguet, Guillaume and Zhang, Yanlei and Rector-Brooks, Jarrid and Fatras, Kilian and Wolf, Guy and Bengio, Yoshua},
  journal={arXiv preprint arXiv:2302.00482},
  year={2023}
}

@inproceedings{goodfellow2014,
  title={Generative adversarial nets},
  author={Goodfellow, Ian and Pouget-Abadie, Jean and Mirza, Mehdi and Xu, Bing and Warde-Farley, David and Ozair, Sherjil and Courville, Aaron and Bengio, Yoshua},
  booktitle={Advances in neural information processing systems},
  pages={2672--2680},
  year={2014}
}

@inproceedings{ronneberger2015u,
  title={U-net: Convolutional networks for biomedical image segmentation},
  author={Ronneberger, Olaf and Fischer, Philipp and Brox, Thomas},
  booktitle={Medical image computing and computer-assisted intervention--MICCAI 2015: 18th international conference, Munich, Germany, October 5-9, 2015, proceedings, part III 18},
  pages={234--241},
  year={2015},
  organization={Springer}
}

@inproceedings{rombach2022high,
  title={High-resolution image synthesis with latent diffusion models},
  author={Rombach, Robin and Blattmann, Andreas and Lorenz, Dominik and Esser, Patrick and Ommer, Bj{\"o}rn},
  booktitle={Proceedings of the IEEE/CVF conference on computer vision and pattern recognition},
  pages={10684--10695},
  year={2022}
}

@article{lu2021learning,
  title={Learning nonlinear operators via {DeepONet} based on the universal approximation theorem of operators},
  author={Lu, Lu and Jin, Pengzhan and Pang, Guofei and Zhang, Zhongqiang and Karniadakis, George Em},
  journal={Nature machine intelligence},
  volume={3},
  number={3},
  pages={218--229},
  year={2021},
  publisher={Nature Publishing Group UK London}
}

@article{wu2020enforcing,
  title={Enforcing statistical constraints in generative adversarial networks for modeling chaotic dynamical systems},
  author={Wu, Jin-Long and Kashinath, Karthik and Albert, Adrian and Chirila, Dragos and Xiao, Heng and others},
  journal={Journal of Computational Physics},
  volume={406},
  pages={109209},
  year={2020},
  publisher={Elsevier}
}

@article{RAISSI2019686,
	author = {M. Raissi and P. Perdikaris and G.E. Karniadakis},
	doi = {https://doi.org/10.1016/j.jcp.2018.10.045},
	issn = {0021-9991},
	journal = {Journal of Computational Physics},
	pages = {686-707},
	title = {Physics-informed neural networks: A deep learning framework for solving forward and inverse problems involving nonlinear partial differential equations},
	url = {https://www.sciencedirect.com/science/article/pii/S0021999118307125},
	volume = {378},
	year = {2019}}

@inproceedings{tran2020gans,
  title={{GAN}s enabled super-resolution reconstruction of wind field},
  author={Tran, Duy Tan and Robinson, Haakon and Rasheed, Adil and San, Omer and Tabib, Mandar and Kvamsdal, Trond},
  booktitle={Journal of Physics: Conference Series},
  volume={1669},
  number={1},
  pages={012029},
  year={2020},
  organization={IOP Publishing}
}

@article{du2024confild,
  title={{CoNFiLD}: Conditional Neural Field Latent Diffusion Model Generating Spatiotemporal Turbulence},
  author={Du, Pan and Parikh, Meet Hemant and Fan, Xiantao and Liu, Xin-Yang and Wang, Jian-Xun},
  journal={arXiv preprint arXiv:2403.05940},
  year={2024}
}

@misc{bastek2025physicsinformeddiffusionmodels,
      title={Physics-Informed Diffusion Models}, 
      author={Jan-Hendrik Bastek and WaiChing Sun and Dennis M. Kochmann},
      year={2025},
      eprint={2403.14404},
      archivePrefix={arXiv},
      primaryClass={cs.LG},
      url={https://arxiv.org/abs/2403.14404}, 
}

@article{xie_2018,
author = {Xie, You and Franz, Erik and Chu, Mengyu and Thuerey, Nils},
title = {tempoGAN: a temporally coherent, volumetric GAN for super-resolution fluid flow},
year = {2018},
issue_date = {August 2018},
publisher = {Association for Computing Machinery},
address = {New York, NY, USA},
volume = {37},
number = {4},
issn = {0730-0301},
url = {https://doi.org/10.1145/3197517.3201304},
doi = {10.1145/3197517.3201304},
journal = {ACM Trans. Graph.},
month = jul,
articleno = {95},
numpages = {15}
}

@misc{christopher2024constrainedsynthesisprojecteddiffusion,
      title={Constrained Synthesis with Projected Diffusion Models}, 
      author={Jacob K Christopher and Stephen Baek and Ferdinando Fioretto},
      year={2024},
      eprint={2402.03559},
      archivePrefix={arXiv},
      primaryClass={cs.LG},
      url={https://arxiv.org/abs/2402.03559}, 
}

@misc{jacobsen2024cocogenphysicallyconsistentconditionedscorebased,
      title={Co{C}o{G}en: Physically-Consistent and Conditioned Score-based Generative Models for Forward and Inverse Problems}, 
      author={Christian Jacobsen and Yilin Zhuang and Karthik Duraisamy},
      year={2024},
      eprint={2312.10527},
      archivePrefix={arXiv},
      primaryClass={cs.LG},
      url={https://arxiv.org/abs/2312.10527}, 
}

@misc{song2024solvinginverseproblemslatent,
      title={Solving Inverse Problems with Latent Diffusion Models via Hard Data Consistency}, 
      author={Bowen Song and Soo Min Kwon and Zecheng Zhang and Xinyu Hu and Qing Qu and Liyue Shen},
      year={2024},
      eprint={2307.08123},
      archivePrefix={arXiv},
      primaryClass={cs.CV},
      url={https://arxiv.org/abs/2307.08123}, 
}

@article{Dasgupta_2025,
   title={Conditional score-based diffusion models for solving inverse elasticity problems},
   volume={433},
   ISSN={0045-7825},
   url={http://dx.doi.org/10.1016/j.cma.2024.117425},
   DOI={10.1016/j.cma.2024.117425},
   journal={Computer Methods in Applied Mechanics and Engineering},
   publisher={Elsevier BV},
   author={Dasgupta, Agnimitra and Ramaswamy, Harisankar and Murgoitio-Esandi, Javier and Foo, Ken Y. and Li, Runze and Zhou, Qifa and Kennedy, Brendan F. and Oberai, Assad A.},
   year={2025},
   month=jan, pages={117425} }

@article{Shu_2023,
   title={A physics-informed diffusion model for high-fidelity flow field reconstruction},
   volume={478},
   ISSN={0021-9991},
   url={http://dx.doi.org/10.1016/j.jcp.2023.111972},
   DOI={10.1016/j.jcp.2023.111972},
   journal={Journal of Computational Physics},
   publisher={Elsevier BV},
   author={Shu, Dule and Li, Zijie and Barati Farimani, Amir},
   year={2023},
   month=apr, pages={111972} }

@misc{huang2024diffusionpdegenerativepdesolvingpartial,
      title={Diffusion{PDE}: Generative {PDE}-Solving Under Partial Observation}, 
      author={Jiahe Huang and Guandao Yang and Zichen Wang and Jeong Joon Park},
      year={2024},
      eprint={2406.17763},
      archivePrefix={arXiv},
      primaryClass={cs.LG},
      url={https://arxiv.org/abs/2406.17763}, 
}

@article{Yang_2019,
   title={Adversarial uncertainty quantification in physics-informed neural networks},
   volume={394},
   ISSN={0021-9991},
   url={http://dx.doi.org/10.1016/j.jcp.2019.05.027},
   DOI={10.1016/j.jcp.2019.05.027},
   journal={Journal of Computational Physics},
   publisher={Elsevier BV},
   author={Yang, Yibo and Perdikaris, Paris},
   year={2019},
   month=oct, pages={136?152} }

@misc{yang2018physicsinformedgenerativeadversarialnetworks,
      title={Physics-Informed Generative Adversarial Networks for Stochastic Differential Equations}, 
      author={Liu Yang and Dongkun Zhang and George Em Karniadakis},
      year={2018},
      eprint={1811.02033},
      archivePrefix={arXiv},
      primaryClass={stat.ML},
      url={https://arxiv.org/abs/1811.02033}, 
}

@inproceedings{Daw_2021, 
   title={{PID-GAN}: A {GAN} Framework based on a Physics-informed Discriminator for Uncertainty Quantification with Physics},
   url={http://dx.doi.org/10.1145/3447548.3467449},
   DOI={10.1145/3447548.3467449},
   booktitle={Proceedings of the 27th ACM SIGKDD Conference on Knowledge Discovery; Data Mining},
   publisher={ACM},
   author={Daw, Arka and Maruf, M. and Karpatne, Anuj},
   year={2021},
   month=aug, collection={KDD ?21} }

@INPROCEEDINGS{rodrigo-bonet_2024,
  author={Rodrigo-Bonet, Esther and Deligiannis, Nikos},
  booktitle={ICASSP 2024 - 2024 IEEE International Conference on Acoustics, Speech and Signal Processing (ICASSP)}, 
  title={Physics-Guided Variational Graph Autoencoder For Air Quality Inference}, 
  year={2024},
  volume={},
  number={},
  pages={6940-6944},
  doi={10.1109/ICASSP48485.2024.10448194}}

@article{Zhong_2023,
   title={{PI-VAE}: Physics-Informed Variational Auto-Encoder for stochastic differential equations},
   volume={403},
   ISSN={0045-7825},
   url={http://dx.doi.org/10.1016/j.cma.2022.115664},
   DOI={10.1016/j.cma.2022.115664},
   journal={Computer Methods in Applied Mechanics and Engineering},
   publisher={Elsevier BV},
   author={Zhong, Weiheng and Meidani, Hadi},
   year={2023},
   month=jan, pages={115664} }

@inproceedings{NEURIPS2023_cd830afc,
	author = {Shmakov, Alexander and Greif, Kevin and Fenton, Michael and Ghosh, Aishik and Baldi, Pierre and Whiteson, Daniel},
	booktitle = {Advances in Neural Information Processing Systems},
	editor = {A. Oh and T. Naumann and A. Globerson and K. Saenko and M. Hardt and S. Levine},
	pages = {65102--65127},
	publisher = {Curran Associates, Inc.},
	title = {End-To-End Latent Variational Diffusion Models for Inverse Problems in High Energy Physics},
	url = {https://proceedings.neurips.cc/paper_files/paper/2023/file/cd830afc6208a346e4ec5caf1b08b4b4-Paper-Conference.pdf},
	volume = {36},
	year = {2023}}

@inproceedings{
holzschuh2024improving,
title={Improving Flow Matching for Posterior Inference with Physics-based Controls},
author={Benjamin Holzschuh and Nils Thuerey},
booktitle={ICML 2024 Workshop on Structured Probabilistic Inference {\&} Generative Modeling},
year={2024},
url={https://openreview.net/forum?id=9HDL3sH61o}
}

@article{CARASSALE2006323,
	author = {Luigi Carassale and Giovanni Solari},
	doi = {https://doi.org/10.1016/j.jweia.2006.01.004},
	issn = {0167-6105},
	journal = {Journal of Wind Engineering and Industrial Aerodynamics},
	note = {The eighth Italian National Conference on Wind Engineering IN-VENTO-2004},
	number = {5},
	pages = {323-339},
	title = {Monte {C}arlo simulation of wind velocity fields on complex structures},
	url = {https://www.sciencedirect.com/science/article/pii/S0167610506000079},
	volume = {94},
	year = {2006}}

@misc{yang2019highlyscalablephysicsinformedganslearning,
      title={Highly-scalable, physics-informed {GAN}s for learning solutions of stochastic {PDE}s}, 
      author={Liu Yang and Sean Treichler and Thorsten Kurth and Keno Fischer and David Barajas-Solano and Josh Romero and Valentin Churavy and Alexandre Tartakovsky and Michael Houston and Prabhat and George Karniadakis},
      year={2019},
      eprint={1910.13444},
      archivePrefix={arXiv},
      primaryClass={physics.comp-ph},
      url={https://arxiv.org/abs/1910.13444}, 
}

@article{karhunen1947lineare,
  title={{\"U}ber lineare Methoden in der Wahrscheinlichkeitsrechnung},
  author={Karhunen, Kari},
  journal={Ann Acad Sci Fennicae},
  volume={37},
  pages={1},
  year={1947}
}

@article{loeve1955probability,
  title={Probability theory: foundations, random sequences},
  author={Loeve, Michel},
  journal={(No Title)},
  year={1955}
}

@book{xiu2010numerical,
  title={Numerical methods for stochastic computations: a spectral method approach},
  author={Xiu, Dongbin},
  year={2010},
  publisher={Princeton University Press}
}

@book{sutton2009image,
  title={Image correlation for shape, motion and deformation measurements: basic concepts, theory and applications},
  author={Sutton, Michael A and Orteu, Jean Jose and Schreier, Hubert},
  year={2009},
  publisher={Springer Science \& Business Media}
}

@article{olivier2020uqpy,
  title={{UQ}py: A general purpose Python package and development environment for uncertainty quantification},
  author={Olivier, Audrey and Giovanis, Dimitris G and Aakash, BS and Chauhan, Mohit and Vandanapu, Lohit and Shields, Michael D},
  journal={Journal of Computational Science},
  volume={47},
  pages={101204},
  year={2020},
  publisher={Elsevier}
}

@article{shinozuka1991simulation,
  title={Simulation of stochastic processes by spectral representation},
  author={Shinozuka, Masanobu and Deodatis, George},
  year={1991}
}

@techreport{coulter2020sonic,
  title={Sonic Detection and Ranging ({SODAR}) Wind Profiler Instrument Handbook},
  author={Coulter, Richard L and Muradyan, Paytsar},
  year={2020},
  institution={DOE Office of Science Atmospheric Radiation Measurement (ARM) Program}
}

@article{DEODATIS2025110522,
	author = {George Deodatis and Michael Shields},
	doi = {https://doi.org/10.1016/j.ress.2024.110522},
	issn = {0951-8320},
	journal = {Reliability Engineering  {\&} System Safety},
	pages = {110522},
	title = {The Spectral Representation Method: A framework for simulation of stochastic processes, fields, and waves},
	url = {https://www.sciencedirect.com/science/article/pii/S0951832024005945},
	volume = {254},
	year = {2025}}

@article{fukunaga1970application,
  title={Application of the {K}arhunen-{L}oeve expansion to feature selection and ordering},
  author={Fukunaga, Keinosuke and Koontz, Warren LG},
  journal={IEEE Transactions on computers},
  volume={100},
  number={4},
  pages={311--318},
  year={1970},
  publisher={IEEE}
}

@book{reddy_2013, place={Cambridge}, edition={2}, title={An Introduction to Continuum Mechanics}, DOI={10.1017/CBO9781139178952}, publisher={Cambridge University Press}, author={Reddy, J. N.}, year={2013}}

@ARTICLE{1161901,
  author={Welch, P.},
  journal={IEEE Transactions on Audio and Electroacoustics}, 
  title={The use of fast Fourier transform for the estimation of power spectra: A method based on time averaging over short, modified periodograms}, 
  year={1967},
  volume={15},
  number={2},
  pages={70-73},
  doi={10.1109/TAU.1967.1161901}}

@misc{warner2020inverseestimationelasticmodulus,
      title={Inverse Estimation of Elastic Modulus Using Physics-Informed Generative Adversarial Networks}, 
      author={James E. Warner and Julian Cuevas and Geoffrey F. Bomarito and Patrick E. Leser and William P. Leser},
      year={2020},
      eprint={2006.05791},
      archivePrefix={arXiv},
      primaryClass={eess.IV},
      url={https://arxiv.org/abs/2006.05791}, 
}

@misc{shah2025generativemodelingmicroweatherwind,
      title={Generative Modeling of Microweather Wind Velocities for Urban Air Mobility}, 
      author={Tristan A. Shah and Michael C. Stanley and James E. Warner},
      year={2025},
      eprint={2503.02690},
      archivePrefix={arXiv},
      primaryClass={cs.CE},
      url={https://arxiv.org/abs/2503.02690}, 
}

@article{ostoja1998random,
  title={Random field models of heterogeneous materials},
  author={Ostoja-Starzewski, Martin},
  journal={International Journal of Solids and Structures},
  volume={35},
  number={19},
  pages={2429--2455},
  year={1998},
  publisher={Elsevier}
}

@article{doi:10.1137/S0036141002409167,
author = {Mikulevicius, R. and Rozovskii, B. L.},
title = {Stochastic {N}avier-{S}tokes Equations for Turbulent Flows},
journal = {SIAM Journal on Mathematical Analysis},
volume = {35},
number = {5},
pages = {1250-1310},
year = {2004},
doi = {10.1137/S0036141002409167},
URL = { https://doi.org/10.1137/S0036141002409167},
}

@article{Guillot_2015,
   title={Statistical paleoclimate reconstructions via {M}arkov random fields},
   volume={9},
   ISSN={1932-6157},
   url={http://dx.doi.org/10.1214/14-AOAS794},
   DOI={10.1214/14-aoas794},
   number={1},
   journal={The Annals of Applied Statistics},
   publisher={Institute of Mathematical Statistics},
   author={Guillot, Dominique and Rajaratnam, Bala and Emile-Geay, Julien},
   year={2015},
   month=mar }

@misc{openai2024gpt4,
  title        = {{GPT-4 Technical Report}},
  author       = {OpenAI and Josh Achiam and Steven Adler and Sandhini Agarwal and others},
  year         = {2024},
  eprint       = {2303.08774},
  archivePrefix= {arXiv},
  primaryClass = {cs.CL},
  url          = {https://arxiv.org/abs/2303.08774}
}

@article{wang2021understanding,
  title={Understanding and mitigating gradient flow pathologies in physics-informed neural networks},
  author={Wang, Sifan and Teng, Yujun and Perdikaris, Paris},
  journal={SIAM Journal on Scientific Computing},
  volume={43},
  number={5},
  pages={A3055--A3081},
  year={2021},
  publisher={SIAM}
}

@article{daw2022rethinking,
  title={Rethinking the importance of sampling in physics-informed neural networks},
  author={Daw, Arka and Bu, Jie and Wang, Sifan and Perdikaris, Paris and Karpatne, Anuj},
  journal={arXiv preprint arXiv:2207.02338},
  year={2022}
}

@article{hou2023enhancing,
  title={Enhancing {PINN}s for solving {PDE}s via adaptive collocation point movement and adaptive loss weighting},
  author={Hou, Jie and Li, Ying and Ying, Shihui},
  journal={Nonlinear Dynamics},
  volume={111},
  number={16},
  pages={15233--15261},
  year={2023},
  publisher={Springer}
}

@misc{zampini_training-free_2025,
	title = {Training-{Free} {Constrained} {Generation} {With} {Stable} {Diffusion} {Models}},
	url = {http://arxiv.org/abs/2502.05625},
	doi = {10.48550/arXiv.2502.05625},
	language = {en},
	urldate = {2025-12-10},
	publisher = {arXiv},
	author = {Zampini, Stefano and Christopher, Jacob K. and Oneto, Luca and Anguita, Davide and Fioretto, Ferdinando},
	month = oct,
	year = {2025},
	note = {arXiv:2502.05625 [cs]},
}

@misc{liang_chance-constrained_2025,
	title = {Chance-constrained {Flow} {Matching} for {High}-{Fidelity} {Constraint}-aware {Generation}},
	url = {http://arxiv.org/abs/2509.25157},
	doi = {10.48550/arXiv.2509.25157},
	language = {en},
	urldate = {2025-12-10},
	publisher = {arXiv},
	author = {Liang, Jinhao and Sun, Yixuan and Samaddar, Anirban and Madireddy, Sandeep and Fioretto, Ferdinando},
	month = sep,
	year = {2025},
	note = {arXiv:2509.25157 [cs]},
}

@misc{utkarsh_physics-constrained_2025,
	title = {Physics-{Constrained} {Flow} {Matching}: {Sampling} {Generative} {Models} with {Hard} {Constraints}},
	shorttitle = {Physics-{Constrained} {Flow} {Matching}},
	url = {http://arxiv.org/abs/2506.04171},
	doi = {10.48550/arXiv.2506.04171},
	language = {en},
	urldate = {2025-12-10},
	publisher = {arXiv},
	author = {Utkarsh, Utkarsh and Cai, Pengfei and Edelman, Alan and Gomez-Bombarelli, Rafael and Rackauckas, Christopher Vincent},
	month = nov,
	year = {2025},
	note = {arXiv:2506.04171 [cs]},
}

@article{WANG2022110768,
	author = {Sifan Wang and Xinling Yu and Paris Perdikaris},
	doi = {https://doi.org/10.1016/j.jcp.2021.110768},
	issn = {0021-9991},
	journal = {Journal of Computational Physics},
	pages = {110768},
	title = {When and why PINNs fail to train: A neural tangent kernel perspective},
	url = {https://www.sciencedirect.com/science/article/pii/S002199912100663X},
	volume = {449},
	year = {2022}}

@article{WANG2024116813,
	author = {Sifan Wang and Shyam Sankaran and Paris Perdikaris},
	doi = {https://doi.org/10.1016/j.cma.2024.116813},
	issn = {0045-7825},
	journal = {Computer Methods in Applied Mechanics and Engineering},
	pages = {116813},
	title = {Respecting causality for training physics-informed neural networks},
	url = {https://www.sciencedirect.com/science/article/pii/S0045782524000690},
	volume = {421},
	year = {2024}}

@article{10.1016/j.neucom.2022.05.015,
author = {Xiang, Zixue and Peng, Wei and Liu, Xu and Yao, Wen},
title = {Self-adaptive loss balanced Physics-informed neural networks},
year = {2022},
issue_date = {Jul 2022},
publisher = {Elsevier Science Publishers B. V.},
address = {NLD},
volume = {496},
number = {C},
issn = {0925-2312},
url = {https://doi.org/10.1016/j.neucom.2022.05.015},
doi = {10.1016/j.neucom.2022.05.015},
journal = {Neurocomput.},
month = jul,
pages = {11?34},
numpages = {24}
}

@article{guo_2019,
	author = {Guo, Zhilin and Brusseau, Mark and Labolle, Eric and Lopez, Jose},
	doi = {10.1007/s10040-019-01938-9},
	journal = {Hydrogeology Journal},
	month = {02},
	title = {Modeling groundwater contaminant transport in the presence of large heterogeneity: A case study comparing MT3D and RWhet},
	volume = {27},
	year = {2019}}

@article{HAGHIGHAT2021113741,
	author = {Ehsan Haghighat and Maziar Raissi and Adrian Moure and Hector Gomez and Ruben Juanes},
	doi = {https://doi.org/10.1016/j.cma.2021.113741},
	issn = {0045-7825},
	journal = {Computer Methods in Applied Mechanics and Engineering},
	pages = {113741},
	title = {A physics-informed deep learning framework for inversion and surrogate modeling in solid mechanics},
	url = {https://www.sciencedirect.com/science/article/pii/S0045782521000773},
	volume = {379},
	year = {2021}}

@article{ZHAO2025107665,
	author = {B.R. Zhao and D.K. Sun and H. Wu and C.J. Qin and Q.G. Fei},
	doi = {https://doi.org/10.1016/j.neunet.2025.107665},
	issn = {0893-6080},
	journal = {Neural Networks},
	pages = {107665},
	title = {Physics-informed neural networks for solving inverse problems in phase field models},
	url = {https://www.sciencedirect.com/science/article/pii/S0893608025005453},
	volume = {190},
	year = {2025}}

\FloatBarrier

\appendix

\section{Appendix}

\subsection{Application Details}

\subsubsection{Wind Velocity Field Estimation}
\label{sec:wind_appendix}

Further information on the random wind velocity field reconstruction application are provided here, focusing in particular on details of the process for generating training data, the implementation of the statistical coherence constraint residual, and additional results that further illustrate the performance of the proposed approach.

\subsubsection*{Training Data Generation}

Wind velocity training data was generated based on the classical spectral representation approach \cite{shinozuka1991simulation} for simulating stochastic processes, using a description of spectral properties borrowed from wind engineering literature \citep{CARASSALE2006323}. Here, the formulation was motivated by the need to model wind loading on large structures like bridges probabilistically, so the statistical properties of the training data used here can be considered realistic in this context. 

As mentioned in Section \ref{sec:windEstimation}, wind velocity is idealized as being comprised of two components via Equation \ref{eq:wind_field}: a deterministic mean function, $\boldsymbol{\mu}_u(\mathbf{x})$ (Equation \eqref{eq:wind_mean}), and random turbulence component, $\mathbf{W}(\mathbf{x}, \omega)$, modeled as a zero-mean Gaussian process. The main task in obtaining training data for $\mathbf{U}(\mathbf{x}, \omega)$ is thus to generate random samples of $\mathbf{W}$ with prescribed statistical properties. Here, we largely follow \cite{CARASSALE2006323} for the formulation of these properties for realistic wind fields, with the simplification of focusing on a single component of velocity, $U_1$, and assuming $\mathbf{W}$ is aligned with axes of the global reference system, removing the need for coordinate transformations. 

The basis of the formulation is the definition of the cross-power spectral density (CPSD) function (the generalization of the covariance function to include a dependence on frequency, $n$) describing $W$,
\begin{equation}
	S^{CP}_{11}(\mathbf{x}, \mathbf{x}', n) = \sqrt{S_{1}(\mathbf{x}; n) S_{1}(\mathbf{x}'; n)} \, \text{Coh}(\mathbf{x}, \mathbf{x}', n) \label{eq:cpsd}
\end{equation}
where $S_{11}$ is the auto spectral density of the $W_1$ turbulence component and $\text{Coh}$ is the coherence function, as defined in Equation \ref{eq:coherence_function}. A suitable definition of $S_{11}$ for modeling turbulence is given as follows
\begin{equation}
	\frac{n S_{11}(\mathbf{x}; n)}{\sigma_{1}^2(\mathbf{x})} = \frac{\lambda_1 n (L_1 / \| \boldsymbol{\mu}_u(\mathbf{x}) \|)}{(1 + 1.5 \lambda_1 n (L_1 / \| \boldsymbol{\mu}_u(\mathbf{x}) \|))^{5/3}}\label{eq:auto_spectra}
\end{equation}
where  $\lambda_1 = 6.868$ and $\sigma_{1}^2$ and $L_1$ are the variance and integral length scale of $W_1$, given by
\begin{equation}
    \begin{aligned}
         &  \sigma_{1}^2 = [6 - 1.1 \text{atan}(\text{ln} z_0 + 1.75) ] u_*^2 , \\
         &  L_1 = 300 \left(\frac{x_3}{200} \right)^{0.67+0.05 \text{ln} z_0}.
    \end{aligned}
\end{equation}

The spectral representation method (SRM) was employed to generate samples of $W_1$ using an open-source implementation in \texttt{UQpy} \citep{olivier2020uqpy}. SRM is a classical framework for simulating random processes and fields that uses the fact that a zero-mean, stationary stochastic process with known CPSD can be approximated by a finite series representation of cosine terms with random phase angles \citep{DEODATIS2025110522}. To apply the SRM, a CPSD matrix, $\mathbf{S}^{CP} \in \mathbf{R}^{N_{x_2} \times N_{x_3} \times N_{f}}$, must be assembled by evaluating the CPSD in Equation \ref{eq:cpsd} on a spatial and frequency discretization. A $10 \times 10$ uniformly spatial grid across $x_2 \in [0, 100]$ m and $x_3 \in [H_{min}, 100]$ m was used, where $H_{min}=1$ to avoid numerical issues due to Equation \ref{eq:wind_mean} for $x_3 \leq z_0$. A grid of $N_f = 128$ uniform points for $f \in [0, 3.0]$ Hz was used to discretize frequency ($\Delta_f = 3.0 / N_f$).  Following \cite{CARASSALE2006323}, this choice of frequency discretization simulates $T=1/\Delta_f = 42.5$ seconds with $N_t = 2 \times N_f = 256$ time steps. Sparse measurement data was simulated for the example by keeping every third vertical column of sensors, starting at $x_2=0$ m, filtering data from $60$ total grid points (see Figure \ref{fig:windExample} a). 
A Python implementation of the above procedure is included in the code repository for the paper. 

\begin{figure}[htbp]
	\centering
	\begin{subfigure}{0.99\textwidth}
		\includegraphics[width=\linewidth]{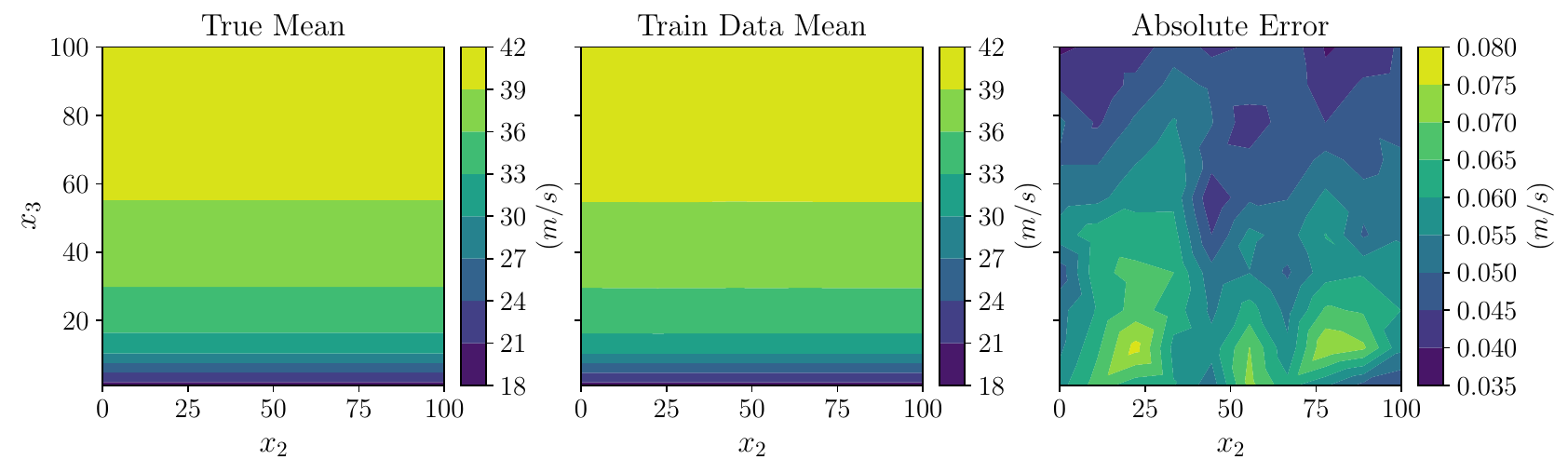}
		\caption{} 
		\label{fig:windSub1}
	\end{subfigure}
	\begin{subfigure}{0.99\textwidth}
		\includegraphics[width=1.05\linewidth]{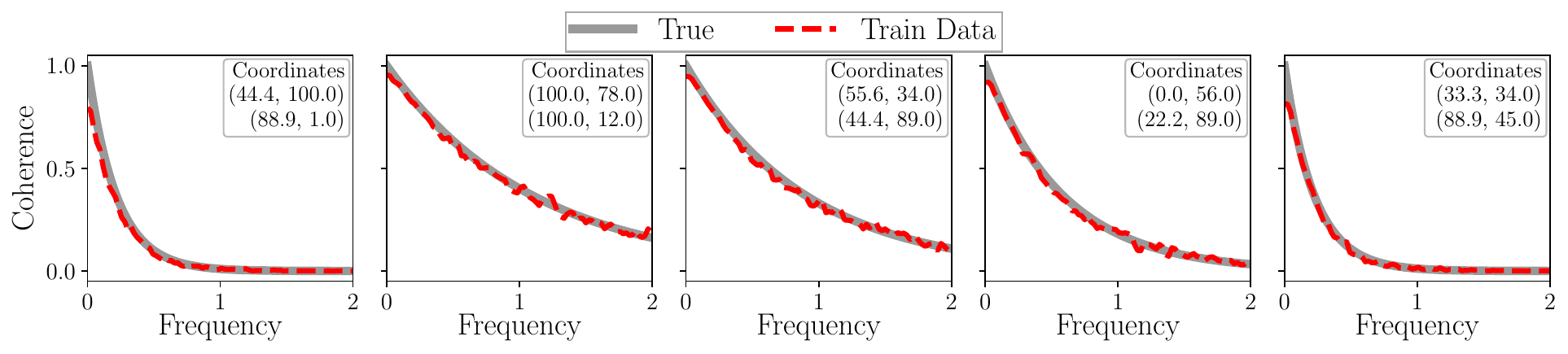}
		\caption{} 
		\label{fig:windSub2}
	\end{subfigure}
	\caption{Wind velocity field training data verification. Comparison of the true and empirically-calculated values from the training data for the (a) mean wind field and (b) coherence between the wind velocity at five random pairs of spatial coordinates.}
	\label{fig:windTrainDataVerification}
\end{figure}

A verification of synthetically-generated wind velocity used for training data is shown in Figure \ref{fig:windTrainDataVerification}. Here, figure (a) compares the mean of the training data to the true mean function in Equation \ref{eq:wind_mean} and (b) compares the empirically-calculated coherence  using the training data (see next section) with the prescribed coherence in Equation \ref{eq:coherence_function}. It can be seen that good agreement is observed between the training data and prescribed values.

\subsubsection*{Residual Evaluation}

The wind field reconstruction example leverages a statistical constraint that relies on the residual between the true coherence function (Equation \ref{eq:coherence_function}) and the empirically estimated value from DGM wind velocity samples, $\hat{u}(\mathbf{x})$. Let the generated wind velocity at a given spatial collocation point, $\tilde{\mathbf{x}}^{(c)} = (\tilde{x}_2^{(c)}, \tilde{x}_3^{(c)})$, be given as $\hat{u}_1^{c} \equiv \hat{u}(\tilde{x}_2^{(c)}, \tilde{x}_3^{(c)}, t)$. The coherence between wind velocities at two different collocation points, $\tilde{\mathbf{x}}^{(c_1)}$ and $\tilde{\mathbf{x}}^{(c_2)}$, can be empirically estimated as:
\begin{equation}
	\gamma^2(f) = \frac{ | P_{12}(f) |^2}{P_{11}(f)  P_{22}(f) },
\end{equation}
where $P_{12}$ is the (empirical) CPSD between $\hat{u}_1^{c_1}$ and $\hat{u}_1^{c_2}$ and $P_{11}$ and $P_{22}$ are the (empirical) auto spectra of $\hat{u}_1^{c_1}$ and  $\hat{u}_1^{c_2}$, respectively. The empirical spectral densities, $P_{**}$, are computed using Welch's method \citep{1161901} based on the Fast Fourier Transform (FFT) in \texttt{PyTorch}. Finally, the coherence residual for a single pair of collocation points is calculated as follows:
\begin{equation}
	\frac{1}{N_f} \sum_{i=1}^{N_f} \| \gamma^2(f^{(i)} ) - \text{Coh}(\tilde{\mathbf{x}}^{(c_1)},\tilde{\mathbf{x}}^{(c_2)}, f^{(i)}) \|^2.
\end{equation}

\subsubsection*{Additional Results}

\begin{figure}[htbp]
	\centering
	\includegraphics[width=\textwidth]{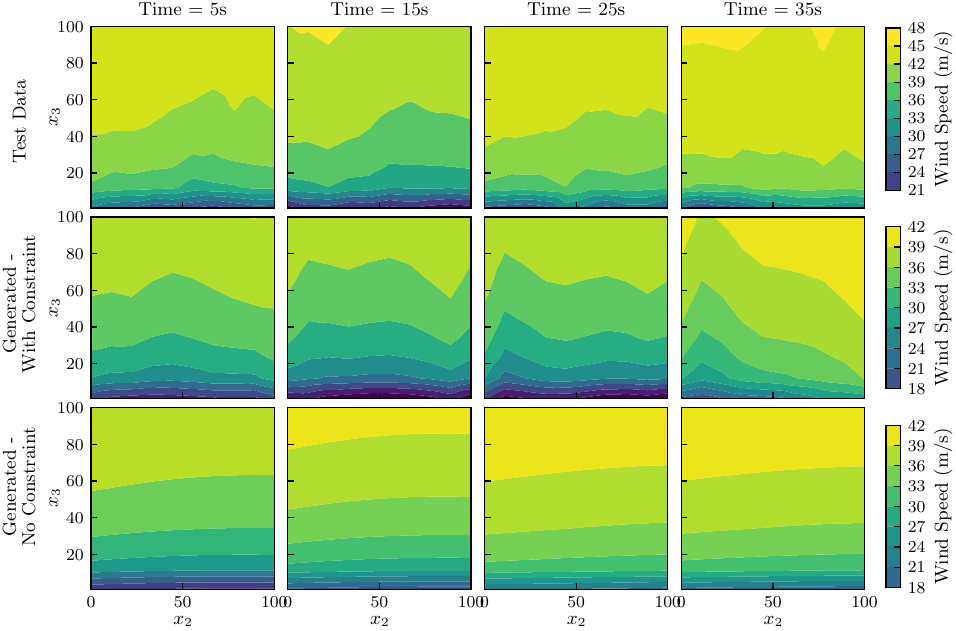}
	\caption{Representative wind velocity samples at four different instances in time for the test data (top), generated with coherence constraint (middle), and generated without coherence constraint (bottom).}
	\label{fig:windSamples}
\end{figure}

Figure \ref{fig:windSamples} compares a representative wind velocity sample from the test dataset versus generated samples both with and without the coherence constraint incorporated during training. Four snapshots in time are shown for each sample. It can be seen that imposing the coherence constraint with c-LFM allows more realistic variability in the horizontal direction to be captured, while the generated sample with no constraint has nearly no variation in the horizontal direction due to the sparseness of the sensors in this direction (only four locations along $x_2$ - see Figure \ref{fig:windExample} (a)).

\begin{figure}[htbp]
	\centering
	\includegraphics[width=\textwidth]{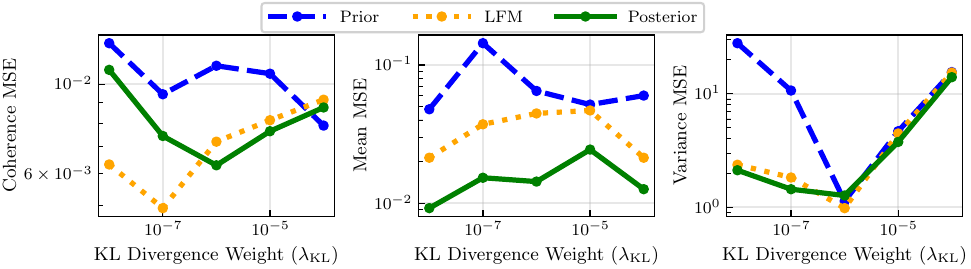}
	\caption{Performance of different sampling approaches for wind velocity estimation in terms of coherence (left), mean (middle), and variance (right) MSE as a function of the KL divergence weight ($\lambda_{KL}$).}
	\label{fig:windErrorsVsKLDWeight}
\end{figure}

Figure \ref{fig:windErrorsVsKLDWeight} shows the performance of c-LFM versus both prior and posterior sampling with a constrained VAE as a function of KL divergence weight, $\lambda_{KL}$. Here, the results are reported for $\lambda_{r} = 0.01$, which was found to effectively balance the statistical errors, and with errors averaged over three random trials. The results generally show a similar ranking in accuracy as found throughout the previous results: posterior sampling showing lowest errors followed by LFM and then prior sampling (constrained VAE). While the coherence and mean MSE versus $\lambda_{KL}$ results lack a well-defined trend, the variance MSE shows the expected behaviors foreshadowed by Figure \ref{fig:latent_sampling}. That is, LFM and posterior sampling show more robustness to choice of $\lambda_{KL}$ relative to prior sampling, where variance error increases rapidly for values larger or smaller than $\lambda_{KL} = 10^{-6}$.

\subsubsection{Material Property Characterization}\label{sec:appendix_details:material}

\begin{figure}[htbp]
    \centering
    \begin{subfigure}{0.49\textwidth}
        \includegraphics[width=\linewidth]{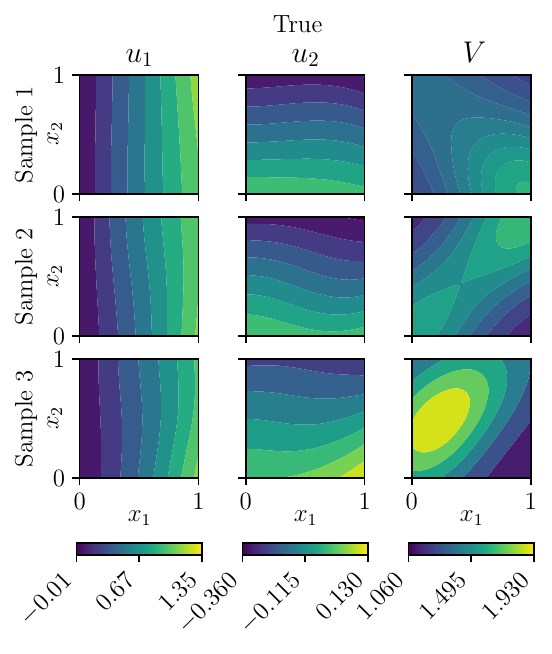}
        \caption{}
    \end{subfigure}
    \begin{subfigure}{0.49\textwidth}
        \includegraphics[width=\linewidth]{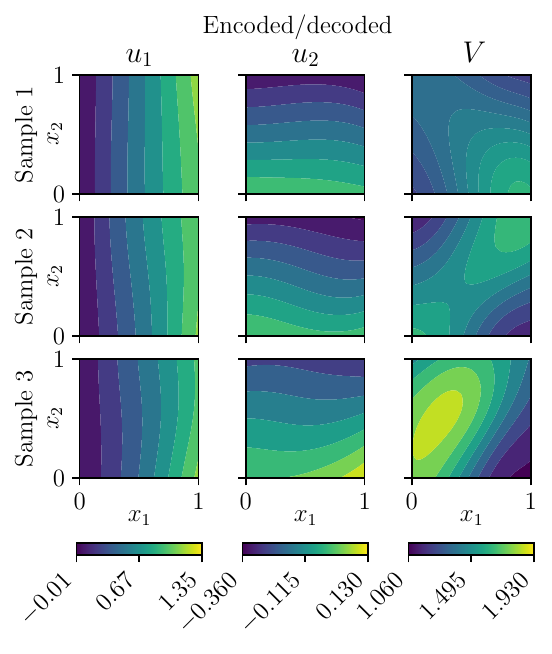}
        \caption{}
    \end{subfigure}
    \caption{(a) Samples from the true $\mathbf{U}$ and $V$ random fields and (b) encoded/decoded versions ($N=1000$) of the same samples.}
    \label{fig:materialVAEResult}
\end{figure}

Additional results for the material property characterization application are provided here. A qualitative assessment of the VAE is presented in Figure \ref{fig:materialVAEResult}. Three ground truth samples are shown in subfigure (a), with the corresponding encoded/decoded ($N=1000$) fields shown in subfigure (b). The encoder and decoder must work in concert to correctly recreate the sampled fields, including correct application of the physical constraints since field samples $V(\mathbf{x}, \hat{\omega})$ are unknown and not provided at training time nor to the encoder in this example. While there is close correspondence between the true and encoded/decoded observations ($u_1$ and $u_2$), there is noticeable discrepancy between samples of the inferred field, $V$, indicating a relative lack of accuracy in the enforcement of the physical constraint. Given the well-known challenges in training neural networks with physics residuals \citep{wang2021understanding}, more advanced training approaches, including adaptive loss weighting \citep{hou2023enhancing} and collocation point sampling \citep{daw2022rethinking}, could improve the performance when using c-LFM with physical constraints.

\begin{figure}[htbp]
    \centering
    \includegraphics[width=0.49\linewidth]{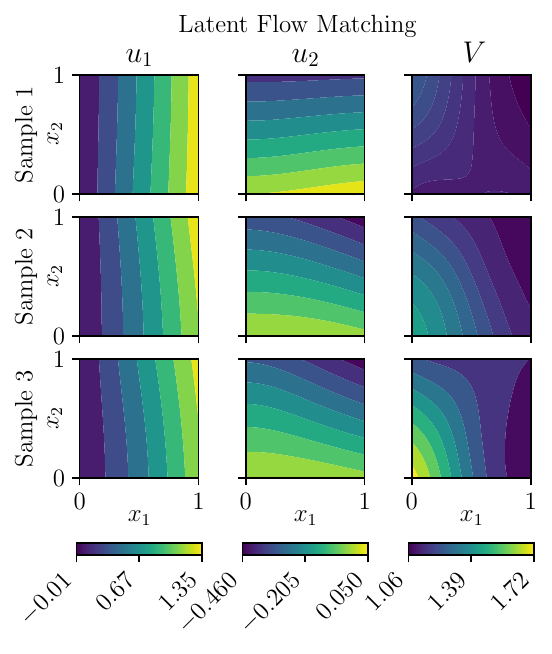}
    \caption{Generated samples from the estimated $\mathbf{U}$ and $V$ random fields ($N=1000$).}
    \label{fig:materialSamples}
\end{figure}

Once trained, LFM allows for the generation of new samples, as illustrated in Figure \ref{fig:materialSamples}. Note that these samples were randomly generated and do not correspond with the samples shown in Figure \ref{fig:materialVAEResult}.

\subsection{Implementation Details}\label{sec:appendix_implementation}

This section provides further implementation details for the proposed approach.  The main training pipeline for the VAE was implemented using PyTorch Lightning with GPU acceleration and model checkpointing based on the minimum training loss. The Adam optimizer was used with $\beta_1=0.9$ and  $\beta_2=0.999$.  All experiments were conducted on an internal computing node equipped with eight NVIDIA A100 GPUs (80GB memory per GPU), 128 CPU cores, and 750GB of system memory.

Network architecture and hyper-parameters used for the VAEs in each of the six examples are summarized in Table \ref{tab:nn_architecture_details}. Most of the networks used were multi-layer perceptrons (MLPs) with GELU activation functions. In the two application examples, it was found that more expressive architectures for the branch and encoder was beneficial for the added problem complexity. For these examples, an enhanced branch network was used with residual blocks containing SiLU activations, layer normalization, and skip connections to improve gradient flow while enabling deeper representations necessary for capturing the more complex function spaces (the number of hidden layers in Table \ref{tab:nn_architecture_details} corresponds to number of residual blocks used). For the wind velocity field estimation example, the encoder portion of the U-Net architecture\citep{ronneberger2015u} was adopted for a more capable encoder. The objective function weights ($\lambda_{kl}$, $\lambda_f$, $\lambda_r$) listed in Table \ref{tab:nn_architecture_details} reflect best-case or default values used in the examples while results were also shown for different weight values throughout Section \ref{sec:experiments}.

\begin{table}[ht]
    \centering
    \caption{Neural network architecture and training details}
    \begin{tabular}{lcccccc}
        \toprule
        & \multicolumn{2}{c}{\textbf{Reconstruction Demos}} & \multicolumn{2}{c}{\textbf{Inference Demos}} & \multicolumn{2}{c}{\textbf{Applications}} \\
        \cmidrule(lr){2-3} \cmidrule(lr){4-5} \cmidrule(lr){6-7}
        \textbf{Section}           & \textbf{4.1.1} & \textbf{4.1.2} & \textbf{4.2.1} & \textbf{4.2.2} & \textbf{4.3} & \textbf{4.4} \\
        \midrule
        Batch Size                 & 256            & 256            & 256            & 256            & 128          & 256          \\
        Learning Rate              & $10^{-3}$      & $10^{-3}$      & $10^{-3}$      & $10^{-3}$      & $5 \times 10^{-4}$ & $10^{-3}$    \\
        Max. Gradient Norm         & 0.5            & 0.5            & 0.5            & 0.5            & 0.5          & $0.01$    \\
        DeepONet dimension ($p$)   & 64             & 64             & 64             & 64             & 512          & 256          \\
        Encoder \# hidden layers   & 3              & 3              & 3              & 3              & UNet         & 3            \\
        Branch \# hidden layers    & 2              & 2              & 2              & 2              & 2            & 2            \\
        Trunk \# hidden layers     & 2              & 2              & 2              & 2              & 2            & 3            \\
        Encoder hidden layer width & 128            & 128            & 128            & 128            & UNet         & 128          \\
        Branch hidden layer width  & 128            & 128            & 128            & 128            & 128          & 128          \\
        Trunk hidden layer width   & 128            & 128            & 128            & 128            & 128          & 128          \\
        $\lambda_{kl}$             & $10^{-6}$      & $10^{-3}$      & $10^{-6}$      & $10^{-6}$      & $10^{-6}$    & $10^{-6}$    \\
        $\lambda_{r}$              & $10^{-2}$      & --             & --             & --             & $10^{-2}$    & --           \\
        $\lambda_{f}$              & --             & $1$            & $10^{-3}$      & $10^{-1}$      & --           & $10^{-3}$    \\
        Epochs                     & $10^4$         & $10^3$         & $10^4$         & $10^4$         & $1.5 \times 10^4$ & $2 \times 10^4$ \\
        \# GPUs                    & 1              & 1              & 1              & 1              & 4            & 1            \\
        Approx. Training Time      & 10 min         & 10 min         & 10 min         & 10 min         & 24 hours     & 6 hours      \\
        \bottomrule
    \end{tabular}
    \label{tab:nn_architecture_details}
\end{table}


For the flow matching implementation used for latent space sampling,  Table~\ref{tab:fm_hyperparams} summarizes the architecture and training hyperparameters used for each example. For the lower-dimensional demonstration problems, a relatively compact MLP with three hidden layers was used as the velocity field model, whereas the higher-dimensional application problems employed a larger network with five hidden layers of 512 units each. All flow models used GELU activation functions. For multimodal random field inference demonstration, an enhanced flow model architecture was employed, which incorporates sinusoidal time embeddings, layer normalization, and residual connections to better capture the multimodal latent distribution.
Training was performed using the Adam optimizer with $\beta_1=0.9$, $\beta_2=0.999$, and gradient clipping with a maximum norm of $1.0$.
Latent codes were normalized (z-scored) prior to flow matching training to improve stability.
During training, a small noise component with standard deviation $\sigma_\mathrm{min}$ was added to the interpolated points.
For sampling, we leveraged the \texttt{torchdyn} library's neural ODE implementation with an adjoint-based Dormand--Prince (\texttt{dopri5}) adaptive solver to integrate the learned velocity field over $t \in [0, 1]$.

\begin{table}[htbp]
    \centering
    \caption{Flow matching hyperparameters for each example. All models use GELU activations and the Adam optimizer ($\beta_1=0.9$, $\beta_2=0.999$) with gradient clipping (max norm $1.0$).}
    \label{tab:fm_hyperparams}
    \begin{tabular}{lcccccc}
        \toprule
        & \multicolumn{2}{c}{\textbf{Reconstruction Demos}} & \multicolumn{2}{c}{\textbf{Inference Demos}} & \multicolumn{2}{c}{\textbf{Applications}} \\
        \cmidrule(lr){2-3} \cmidrule(lr){4-5} \cmidrule(lr){6-7}
        \textbf{Section}  & \textbf{4.1.1} & \textbf{4.1.2} & \textbf{4.2.1} & \textbf{4.2.2} & \textbf{4.3} & \textbf{4.4} \\
        \midrule
        Flow model        & MLP            & MLP            & MLP            & Enhanced$^*$   & MLP          & MLP          \\
        Hidden layers     & 3              & 3              & 3              & 3              & 5            & 5            \\
        Hidden layer size & 128            & 128            & 32             & 128            & 512          & 512          \\
        Learning rate     & $10^{-4}$      & $10^{-4}$      & $10^{-3}$      & $10^{-4}$      & $10^{-4}$    & $10^{-4}$    \\
        Batch size        & 256            & 256            & 256            & 256            & 128          & 512          \\
        $\sigma_\mathrm{min}$ & $10^{-2}$  & $10^{-2}$      & $10^{-3}$      & $10^{-2}$      & $10^{-4}$    & $10^{-4}$    \\
        Epochs            & 500            & 500            & 500            & 1000           & 1000         & 4000         \\
        \bottomrule
        \multicolumn{7}{l}{\footnotesize $^*$Enhanced model uses sinusoidal time embeddings, layer normalization, and residual connections.}
    \end{tabular}
\end{table}

Finally, we acknowledge the adoption of portions of the following open-source resources for the development of code for this work:

\textbf{The Annotated Diffusion Model}\\
\textit{Authors}: Niels Rogge and Kashif Rasul\\
\textit{Source}: \url{https://huggingface.co/blog/annotated-diffusion}\\
\textit{Components Used}: U-Net architecture implementation\\
\textit{License}: N/A\\
\textit{Notes}: We adapted the U-Net architecture with modifications to suit our specific requirements for function decoders.

\textbf{Conditional Flow Matching}\\
\textit{Authors}: Alexander Tong and Kilian Fatras\\
\textit{Source}: \url{https://github.com/atong01/conditional-flow-matching}\\
\textit{Components Used}: PyTorch wrapper for integrating flow matching with TorchDyn\\
\textit{License}: MIT\\
\textit{Notes}: The wrapper was used to interface our flow model with the Neural ODE solver for sampling from the learned flow.

\end{document}